\documentclass[times, review, 10pt]{elsarticle}

\usepackage{amssymb}

\usepackage{lineno}
\usepackage{multirow}
\usepackage{bbm} 
\usepackage{soul}
\usepackage{amsmath}
\usepackage{color}
\usepackage{hyperref}

\usepackage{setspace}
\journal{Pattern Recognition}

\begin{document}




\label{sec:Titlepage}

\begin{frontmatter}

\title{Enhancing Learnable Descriptive Convolutional Vision
Transformer for Face Anti-Spoofing  \tnoteref{t1}}

\author[1]{Pei-Kai Huang\fnref{fn1}}
\ead{alwayswithme@gapp.nthu.edu.tw}

\author[1]{Jun-Xiong Chong}
\ead{jxchong@gapp.nthu.edu.tw}

\author[1]{Ming-Tsung Hsu}
\ead{xmc510063@gapp.nthu.edu.tw}

\author[1]{Fang-Yu Hsu}
\ead{shellyhsu@gapp.nthu.edu.tw}

\author[1]{Chiou-Ting Hsu\corref{cor1}}
\ead{cthsu@cs.nthu.edu.tw}

\cortext[cor1]{Corresponding author}
\fntext[fn1]{This is the first author.}

\affiliation[1]{organization={Department of Computer Science, 
                              National Tsing Hua University},
                addressline={Kuang-Fu Road}, 
                city={Hsinchu},
                postcode={30013},
                country={Taiwan}}







\begin{abstract}
Face anti-spoofing (FAS) heavily relies on identifying live/spoof discriminative features to counter face presentation attacks.  
Recently, we proposed LDCformer to successfully incorporate the Learnable Descriptive Convolution (LDC) into ViT, to model long-range dependency of locally descriptive features for FAS.  
In this paper, we propose three novel training strategies to effectively enhance the training of LDCformer to largely boost its feature characterization capability. 
The first strategy, dual-attention supervision, is developed to learn fine-grained liveness features guided by regional live/spoof attentions.
The second strategy, self-challenging supervision, is designed to enhance the discriminability of the features by generating challenging training data.  
In addition, we propose a third training strategy, transitional triplet mining strategy, through narrowing the cross-domain gap while maintaining the transitional relationship between live and spoof features, to enlarge the domain-generalization capability of LDCformer. 
Extensive experiments show that LDCformer under joint supervision of the three novel training strategies outperforms previous methods. 
\end{abstract}


 
\end{frontmatter}


\section{Introduction} 

Facial recognition and identification systems have facilitated many applications in daily life, such as unlocking cellphones, authenticating mobile payment, and on-line banking. However, incorporation of facial recognition in these applications also incurs potential security risks and requires specific techniques to support the application security. 
Therefore, many face anti-spoofing (FAS) methods have been developed to counter face presentation attacks. 
For example, in \cite{huang2022learnable}, we have developed a Learnable Descriptive Convolution (LDC) by incorporating a learnable local descriptor into vanilla convolutions to boost its representation capacity.
Encouraged by the excellent performance of LDC \cite{huang2022learnable}, in \cite{huang2023ldcformer}, we further developed a Learnable Descriptive Convolutional Vision Transformer (LDCformer) by incorporating LDC features into ViT backbone for modeling long-range and distinguishing characteristics of FAS.

\begin{figure} 
    \begin{minipage}[b]{1.0\linewidth}
      \centering{ 
      \includegraphics[width=12cm]{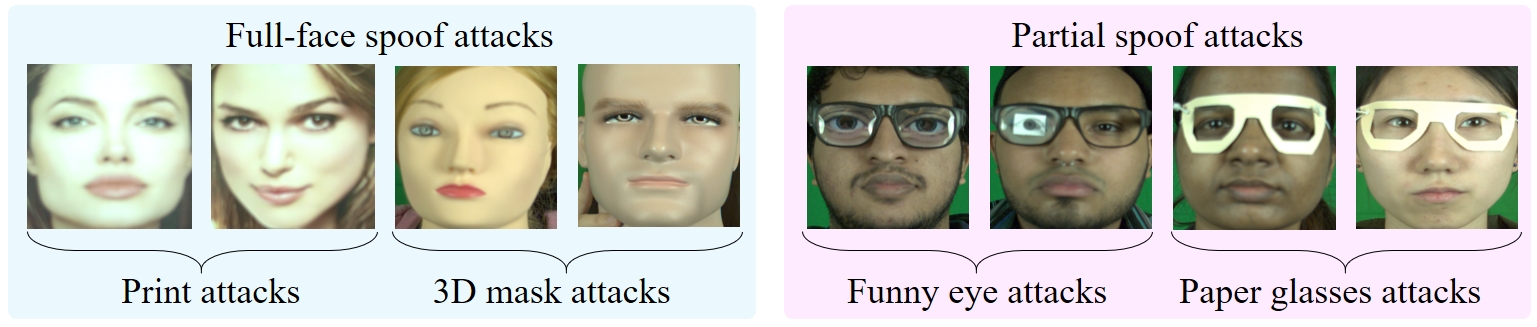}
      }
    \end{minipage}   
\caption{ 
\textcolor{red}{Examples of full-face and partial spoof attacks from the PADISI-Face dataset \cite{rostami2021detection}.}
} 
 \label{fig:partial spoof attacks}   
\end{figure}

Since we simply employ the cross-entropy loss to constrain LDCformer in  \cite{huang2023ldcformer}, \textcolor{red}{there remain three difficult challenges for LDCformer in FAS.
First, as noted in \cite{huang2022learnable}, FAS deals with highly similar characteristics between live and spoof faces and requires a more delicate  representation to accurately characterize the intrinsic features related to face spoofing attacks. 
Hence, the first challenge stems from the lack of fine-grained ground-truth labels to learn fine-grained liveness features.}   
\textcolor{red}{In particular, because} most benchmark datasets provide only binary ground-truth labels to indicate whether an image is live or spoof but give no regional information about where the spoofed regions are located, many methods resorted to auxiliary supervision, e.g., facial depth \cite{huang2022face}, and rPPG signal \cite{huang2022face}, to guide the FAS models on learning fine-grained or physiological  features. 
These auxiliary supervisions, though effective on specific scenarios, heavily rely on the availability and quality of the adopted information and do not generally apply to all the scenarios. 
\textcolor{red}{
For example, facial depth is considered ineffective in detecting 3D mask attacks, because both spoof and live faces exhibit similar facial depth characteristics \cite{yu2020fas}.
As to rPPG signals, because rPPG estimation is highly sensitive to environmental variations, the reliability of using rPPG as auxiliary information in cross-domain scenarios is questionable.
The second challenge in FAS involves detecting subtle partial spoofing attacks. 
While most existing FAS methods are designed to detect full-face spoof attacks, they often overlook the possibility of partial spoof attacks. As shown in Figure~\ref{fig:partial spoof attacks}, partial spoof attacks, such as ``funny eye'' and ``paper glasses'' attacks from the PADISI-Face dataset \cite{rostami2021detection}, target specific facial regions and thus complicate the FAS task on accurately detecting these types of spoof attacks. 
Finally, the third challenge in FAS concerns cross-domain issues.}  
Because different benchmark datasets are independently collected and exhibit various distributions, a model trained on one dataset (i.e., the training domain) usually fails to detect the attacks in other unseen datasets (i.e., unseen domains). 

In this paper, we aim to largely enhance the capabilities of LDCformer \cite{huang2023ldcformer} in FAS by addressing the above \textcolor{red}{three} challenges.
We propose adopting three novel training strategies to jointly supervise the training of LDCformer, including dual-attention supervision, self-challenging supervision, and transitional triplet mining.
First, to address the lack of fine-grained supervision, we introduce a dual-attention supervision by including two jointly trained attention estimators to guide LDCformer  to focus on regional live/spoof attentions for learning fine-grained liveness features.
Next, \textcolor{red}{
to detect subtle partial spoofing attacks, we propose a self-challenging supervision, by generating challenging augmentation data through mixing live and spoof images, to encourage LDCformer to enhance its discriminability between liveness features and partial spoof attacks.} 
Furthermore, to address the cross-domain issue, we propose a transitional triplet mining strategy, by simultaneously narrowing the cross-domain gap and maintaining the transitional relationship between live and spoof features, to enlarge the domain-generalization capability of the learned features.  
These three strategies are developed to collaboratively train LDCformer towards learning highly live/spoof discriminative and domain-generalizable features.
Experimental results on FAS benchmark demonstrate that, under joint supervision of the three training strategies, LDCformer achieves state-of-the-art performance compared to recent face anti-spoofing methods under intra-domain and cross-domain testing scenarios.
 
Our contributions are summarized as follows: 

\noindent 
$\bullet$  
We propose three novel training strategies to cooperatively supervise the training of our previously proposed Learnable Descriptive Convolutional Vision Transformer (LDCformer) \cite{huang2023ldcformer} for largely expanding its capacity 
on modeling long-range and highly distinguishing characteristics of FAS. 

\noindent 
$\bullet$  
\textcolor{red}{
The three novel strategies, including dual-attention supervision, self-challenging supervision, and transitional triplet mining, are developed to explicitly address the lack of fine-grained labels,  enhance detection of subtle partial spoofing attacks, and tackle cross-domain issues.}
Our ablation studies and experimental comparisons all verify their effectiveness on encouraging LDCformer to learn highly discriminative and domain-generalized features. 


\section{Related Work}

\subsection{CNN-based Face Anti-spoofing}

With the huge success of convolutional neural networks (CNNs) in many computer vision tasks, CNN-based methods  have emerged as a favorite for face anti-spoofing. 
Early CNN-based methods \cite{liu2018learning, yu2020fas, shao2020regularized} have demonstrated promising detection performance under intra-domain testing scenarios. 
However, when encountering unseen types of face presentation attack from new domains, these methods usually fail to distinguish the unseen attacks that differ significantly from their training data. 
\textcolor{red}{
To address the issue of poor cross-domain generalization ability, many recent face anti-spoofing methods \cite{huang2022learning, wang2022domain, sun2023rethinking} adopt domain generalization (DG) scenario to learn domain-generalized features to improve the detection performance under cross-domain testing scenarios. 
}

\subsection{ViT-based Face Anti-spoofing}
 
In comparison with CNNs,  vision transformer (ViT) has been proposed to model long-range pixel dependencies, in terms of self-attention (SA) \cite{vaswani2017attention} and multi-head self-attention (MSA)  \cite{vaswani2017attention}, and has achieved significant performance gains over CNNs in many tasks. 
Some recent face anti-spoofing methods \cite{george2021effectiveness,huang2022adaptive,wang2022face}  also relied on  transformers to capture  global semantic information for distinguishing live and spoof faces. In \cite{george2021effectiveness}, the authors proposed to adopt ViT to model long-range data dependencies for face anti-spoofing.
In \cite{huang2022adaptive}, the authors focused on enlarging the diversity of liveness features obtained by ViT to learn generalized liveness features under the few-shot scenario.  
However, because these methods \cite{george2021effectiveness,huang2022adaptive} only adopt binary live/spoof ground-truth labels to supervise ViT, there involves no fine-grained information in guiding ViT towards learning live/spoof discriminative features. 
Therefore, the authors in \cite{wang2022face,wang2022learning} proposed adopting depth supervision to cooperate with ViT for distinguishing live faces from spoof faces.  
However,  facial depth supervision becomes invalid for detecting spoofing attacks concerning no depth difference. 
In addition, although some previous methods \cite{xiao2021early} incorporated vanilla convolution operations into ViT to enable ViT for mobile device and to facilitate efficient training with improved performance \cite{xiao2021early}, explicit analysis on combining the strengths of CNN and ViT still needs further exploration.
Since some CNN-based methods \cite{wang2020deep,yu2020fas,yu2021dual} and our previous method \cite{huang2022learnable} all strive to enhance feature representation of CNN by incorporating different local descriptors into convolution, we believe that incorporation of local feature description is equally essential for ViT  towards learning distinguishing characteristics of FAS.  
Therefore, in \cite{huang2023ldcformer}, we developed an innovative LDCformer by incorporating the Learnable Descriptive Convolution (LDC) \cite{huang2022learnable} into ViT to model long-range and distinguishing characteristics for FAS. 
Since our main focus in \cite{huang2023ldcformer} is on the model design, we simply adopt the cross-entropy loss in terms of binary live/spoof ground truth labels to train the LDCformer but did not focus on the lack of fine-grained supervision and cross-domain challenges.
Therefore, in this paper, to explicitly address these two challenges, 
 we specifically design three novel strategies to cooperatively supervise LDCformer towards learning discriminative and domain-generalizable features for face anti-spoofing. In Section \ref{sec:Proposed Method}, we will present these strategies and explain how we include these strategies to guide the learning of LDCformer. 

\section{Proposed Method}
\label{sec:Proposed Method} 
   
In Section \ref{sec:Learnable Descriptive Convolutional Vision Transformer}, we first revisit the framework of Learnable Descriptive Convolutional Vision Transformer (LDCformer) \cite{huang2023ldcformer}. 
Next, in Section \ref{sec:Training Strategies}, \textcolor{red}{
we present three novel training strategies to collaboratively guide LDCformer in explicitly addressing the lack of fine-grained labels, detecting subtle partial spoofing attacks, and tackling cross-domain testing issues that remain unexplored in \cite{huang2023ldcformer}.
}

\begin{figure} 
    \begin{minipage}[]{1.0\linewidth}
      \centering{ 
      \includegraphics[width=12cm]{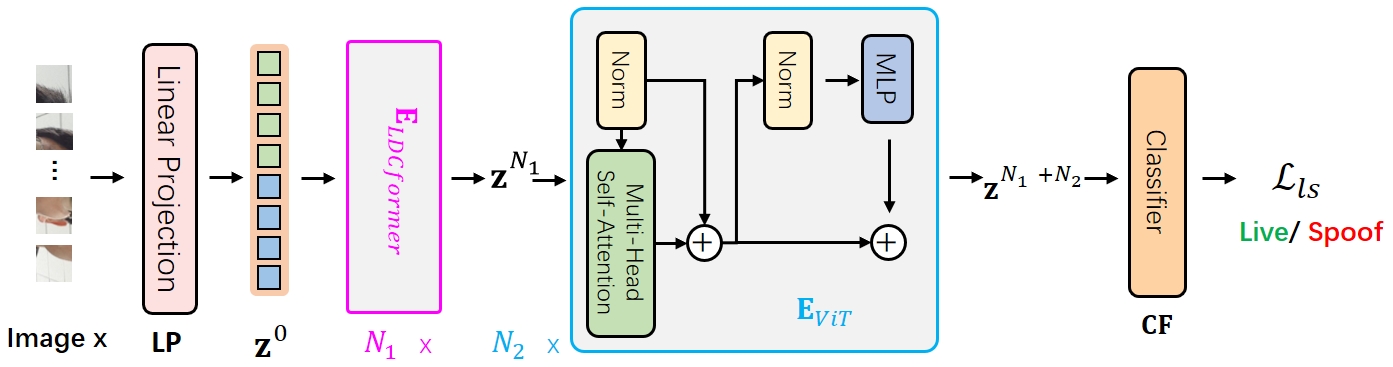}
      }
    \end{minipage}   
\caption{ 
Overview of LDCformer \cite{huang2023ldcformer}, where we additionally include $N_1$ LDCformer encoders $\textbf{E}_{LDCformer}$ (purple solid box) between the linear projection $\textbf{LP}$ and the standard Transformer encoders $\textbf{E}_{ViT}$ (blue solid box) in ViT. 
}
 \label{fig:LDCformer}   
\end{figure}

 \subsection{Learnable Descriptive Convolutional Vision Transformer }
\label{sec:Learnable Descriptive Convolutional Vision Transformer}

Since our previously proposed LDC \cite{huang2022learnable} has been shown to effectively enhance vanilla convolution, in \cite{huang2023ldcformer}, we consider LDC is an ideal candidate to cooperate with ViT and proposed a novel Learnable Descriptive Convolutional Vision Transformer (LDCformer) to facilitate ViT on modeling long-range and locally distinguishing characteristics of FAS. 
 
Figure \ref{fig:LDCformer} gives an overview of LDCformer \cite{huang2023ldcformer}, which consists of a linear projection layer \textbf{LP}, \emph{$N_1$}  LDCformer encoders $\textbf{E}_{LDCformer}$, \emph{$N_2$} standard Transformer encoders $\textbf{E}_{ViT}$, and a live/spoof classifier \textbf{CF}. 
To recap the design of LDCformer \cite{huang2023ldcformer}, below we will first present the patch embeddings, then present the implementation of Multi-Head Self Attention in ViT using the LDCformer Encoder $\mathbf{E}_{LDCformer}$, and finally describe how we use LDCformer to obtain the output features. 

\subsubsection{\textcolor{red}{Patch Embeddings}} 
\label{sec:Patch Embeddings}
First, given an input image $x$ $\in \mathbb{R}^{H\times W\times C}$ of size $H \times W$ with $C $ channels, we use the linear projection $\textbf{LP} \in \mathbb{R}^{P^2 \cdot C \times DN}$ to obtain its patch embedding $\textbf{z}^0$ by
$\textbf{z}^0 = \textbf{LP}(x)  + \textbf{LP}_{pos}$, where $\textbf{LP}_{pos} \in \mathbb{R}^{N \times DN} $ is the position embedding, $P$ is the patch size, $N = HW/P^2$ is the number of patches, and $DN$ is the  number of output channel.

\subsubsection{\textcolor{red}{LDCformer Encoder: LDC-Based Multi-Head Self Attention}} 
\label{sec:LDCformer Encoder}
\begin{figure} 
    \centering{ 
      \includegraphics[height=3.5cm]{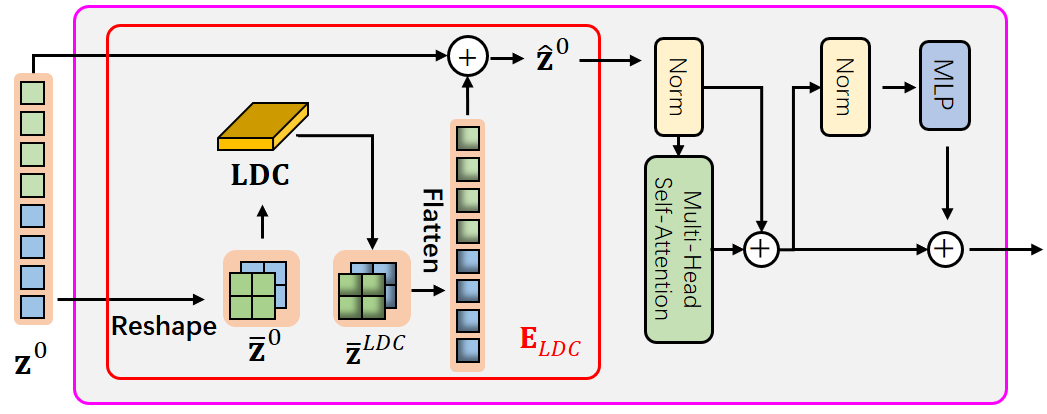} 
      } 
\caption{ 
The LDCformer Encoder $\textbf{E}_{LDCformer}$ in \cite{huang2023ldcformer} (i.e., the purple solid box in Figure \ref{fig:LDCformer}) consists of a LDC encoder $\textbf{E}_{LDC}$ (red solid box), a  LDC-based Multi-Head Self Attention (LDC-MSA), and a multilayer perceptron block (MLP). 
}
 \label{fig:LDCformer_encoder}   
\end{figure}

Next,  we proposed an LDC-based Multi-Head Self Attention (LDC-MSA) by integrating the LDC features into ViT to enhance the representation capability of Multi-Head Self Attention (MSA). 
To implement LDC-MSA, we designed the LDCformer encoder $\textbf{E}_{LDCformer}$, which consists of a LDC encoder $\textbf{E}_{LDC}$, a LDC-MSA, and a multilayer perceptron block (MLP), as shown in Figure \ref{fig:LDCformer_encoder}. 
In particular, in $\textbf{E}_{LDC}$, we reshape $\textbf{z}^0$ into $\Bar{\textbf{z}}^0  \in \mathbb{R}^{D \times P \times P}$ and extract its LDC features, which characterize distinctive local spoofing cues, by $\Bar{\textbf{z}}^{LDC} = \text{LDC}(\Bar{\textbf{z}}^0)$, where $\text{LDC}(\cdot)$ is a single layer LDC \cite{huang2022learnable}, and 
$\Bar{\textbf{z}}^{LDC}$ and $\Bar{\textbf{z}}^0$ are of the same dimension.
Subsequently, we flatten $\Bar{\textbf{z}}^{LDC}$ into ${\textbf{z}}^{LDC}$ and then fuse ${\textbf{z}}^{LDC}$ and ${\textbf{z}}^0$ to obtain the fused feature $\hat{\textbf{z}}^0$ by  $\hat{\textbf{z}}^0  = {\textbf{z}}^0 + \alpha {\textbf{z}}^{LDC}$, where $\alpha$ is a weight factor and is empirically set as $\alpha = 0.15$ in all our experiments.
With $\hat{\textbf{z}}^0$, we involve different self-attention functions by projecting $\hat{\textbf{z}}^0$ in parallel by $\text{LDC-MSA}(\textbf{z}) = [\text{SA}_1(\hat{\textbf{z}}^0),\cdots,\text{SA}_k(\hat{\textbf{z}}^0)] \; \textbf{U}_{ldc-msa}$, 
\noindent
where $\textbf{U}_{ldc-msa}$ is a linear layer, $\text{SA}(\cdot)$ is the standard self-attention function, and $k$ is the number of $\text{SA}(\cdot)$.

\subsubsection{\textcolor{red}{LDCformer}} 
To sum up, as shown in Figure \ref{fig:LDCformer},  we first obtain the patch embedding $\textbf{z}^0$ of the input \textbf{x} and then pass  $\textbf{z}^0$ through $N_1$ $\textbf{E}_{LDCformer}$ to obtain $\textbf{z}^{N_1}$ by $\Tilde{\textbf{z}}^l  = \text{LDC-MSA}(\text{LN}(\hat{\textbf{z}}^{l-1}))+\hat{\textbf{z}}^{l-1}$ and $\textbf{z}^{l} = \text{MLP}(\text{LN}(\Tilde{\textbf{z}}^l))+\Tilde{\textbf{z}}^l$ , where $\text{LN}(\cdot)$ is the LayerNorm function, 
$\text{MLP}(\cdot)$ is the multilayer perceptron block, and $l \in \{1, ..., N_1 \}$.  
Furthermore, we employ $N_2$ standard Transformer encoders $\textbf{E}_{ViT}$ to obtain the output $\textbf{z}^{N_1 + N_2}$ by $\Tilde{\textbf{z}}^l   = \text{MSA}(\text{LN}({\textbf{z}}^{l-1})) $ and $\textbf{z}^{l}  = \text{MLP}(\text{LN}(\Tilde{\textbf{z}}^l))+\Tilde{\textbf{z}}^l$, where $\text{MSA}(\cdot)$ is the multi-head self attention function and
$l \in \{N_1 + 1, N_1 + 2, ..., N_1 + N_2 \}$. 

Finally, we define the liveness loss $\mathcal{L}_{ls}$ to constrain LDCformer by,  
\begin{eqnarray}  
\mathcal{L}_{ls} =
- \sum_{\forall x} y log (\mathbf{CF}(\textbf{z}^{ N_1 + N_2})) + (1-y)log (1-\mathbf{CF}(\textbf{z}^{ N_1 + N_2})), 
\label{eq:liveness_loss}
\end{eqnarray}

\noindent  
where \textbf{CF} is the classifier, and $y$ is the binary liveness label, i.e., $y = 1$ for live images and $y = 0$ for spoof images.

\begin{figure*}[t]
    \begin{minipage}[ ]{1.0\linewidth}
      \centering{ 
      \includegraphics[width=12cm]{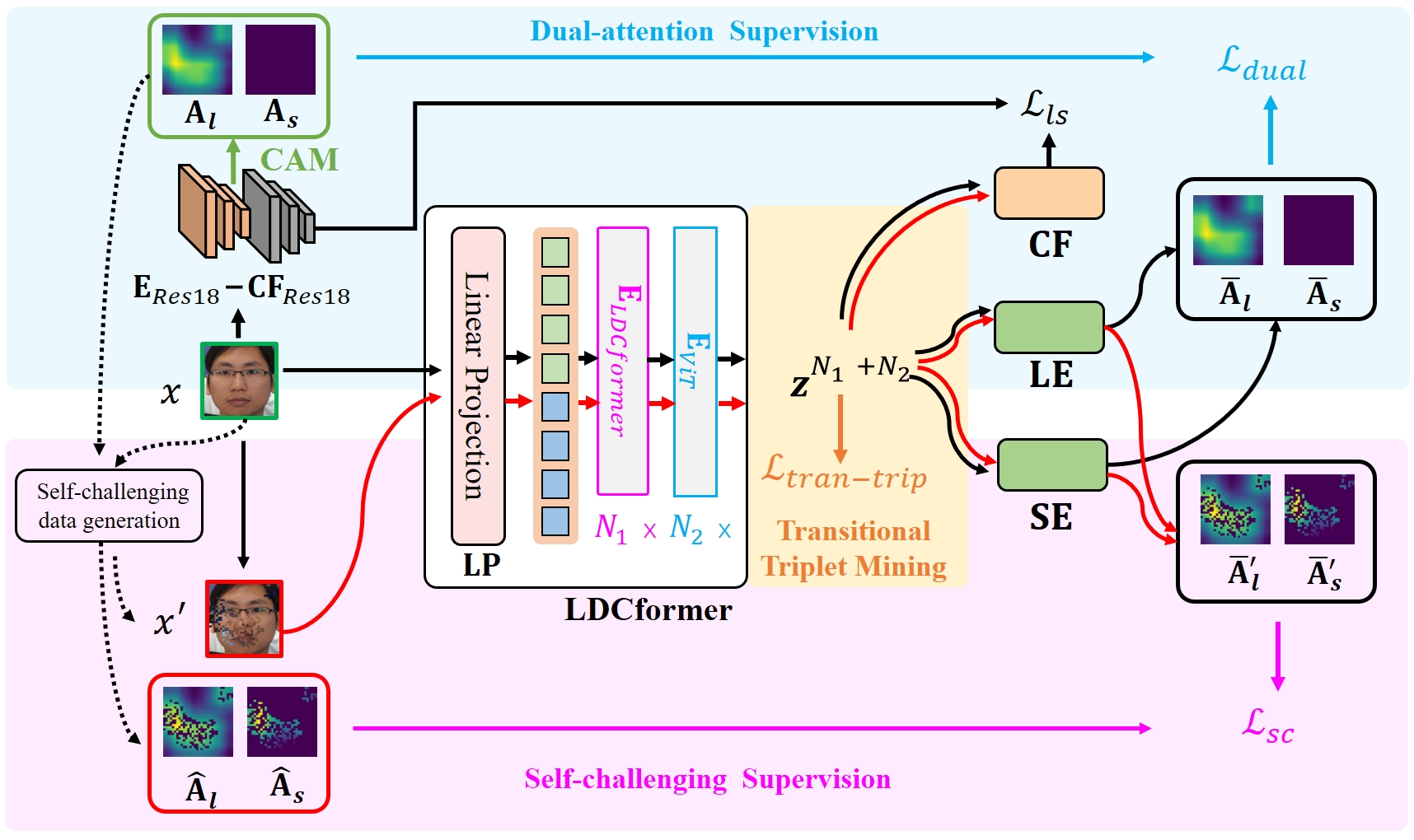} 
      }
    \end{minipage}   
\caption{  
\textcolor{red}{
Illustration  of the proposed three training strategies.
First, to address the lack of fine-grained labels, we adopt the  auxiliary network $\textbf{E}_{Res18}-\textbf{CF}_{Res18}$ to generate the activation maps  $\mathbf{A}_l$ and  $\mathbf{A}_s$, which provide fine-grained supervision for LDCformer (Dual-Attention Supervision $\mathcal{L}_{dual}$).
Next, we mix live and spoof images to generate challenging  images $x^\prime$ and their corresponding attention-based ground truth ${\mathbf{\hat{A}}}_l$ and ${\mathbf{\hat{A}}}_s$ to supervise LDCformer in detecting subtle partial spoofing attacks (Self-Challenging Supervision $\mathcal{L}_{sc}$).
Finally, we examine the relationship between different spoof attacks and live images within the features  $\mathbf{z}^{N1+N2}$ to enhance LDCformer for tackling cross-domain testing issues (Transitional Triplet Mining $\mathcal{L}_{tran-trip}$). 
}} 
 \label{fig:overview}  
\end{figure*}

\subsection{Novel Training Strategies}
\label{sec:Training Strategies} 
 
In \cite{huang2023ldcformer}, we simply adopt the cross-entropy loss in terms of binary live/spoof ground truth labels to train the LCDformer but did not focus on the lack of fine-grained supervision and cross-domain challenges.
Therefore, \textcolor{red}{
 we propose three new training strategies: Dual-Attention Supervision, Self-Challenging Supervision, and Transitional Triplet Mining, to enhance the training of LDCformer. 
}

Figure \ref{fig:overview} shows the overview of the proposed method. In addition to LDCformer \cite{huang2023ldcformer}, we additionally develop an auxiliary network $\textbf{E}_{Res18}-\textbf{CF}_{Res18}$ and two attention estimators \textbf{LE} and \textbf{SE} to facilitate the proposed dual-attention supervision and self-challenging supervision.  
In Figure \ref{fig:overview}, $x \in X$ denotes an input image from the training set $ X = X_l \cup X_s$,  where $X_l$ is the set of live training images, and $X_s = X_p \cup X_r$ is the set of spoof training images consisting of the set of print attack $X_p$ and the set of replay attack $X_r$.
 $x^\prime \in  X^\prime$ denotes an image from the self-challenging dataset  $ X^\prime = X^\prime_p \cup X^\prime_r$, where   $X^\prime_p$ and  $X^\prime_r$  are the sets of mixed-print attack and mixed-replay attack images, respectively. 

 
\subsubsection{\textcolor{red}{Dual-Attention Supervision}}
\label{sec:Dual-Attention Supervision} 
 In addition to the binary live/spoof ground truth labels, we propose including a dual-attention supervision to guide LDCformer on learning fine-grained features. As shown in Figure \ref{fig:overview}, we add a live attention estimator \textbf{LE} and a spoof attention estimator \textbf{SE} after LDCformer to estimate the live and spoof attentions $\Bar{\mathbf{A}}_l$ and $\Bar{\mathbf{A}}_s$. 
In addition, to enforce the two attention estimators \textbf{LE} and \textbf{SE} to focus on class-specific and regional attentions, we additionally develop an auxiliary network $\textbf{E}_{Res18}-\textbf{CF}_{Res18}$ and use its Class Activation Map to generate the quasi-ground truth for \textbf{LE} and \textbf{SE}. Note that, unlike the dual-attention supervision introduced in \cite{huang2022learnable}, where the auxiliary network $\textbf{E}_{Res18}-\textbf{CF}_{Res18}$ is pre-trained and fixed, here we conduct an end-to-end training process to jointly train the auxiliary network $\textbf{E}_{Res18}-\textbf{CF}_{Res18}$ along with LDCformer. This joint training strategy not only simplifies the training pipeline of LDCformer but also effectively enhances the overall performance.

\begin{figure} 
     \begin{tabular}{cc} 
    {\includegraphics[width=5.5cm]{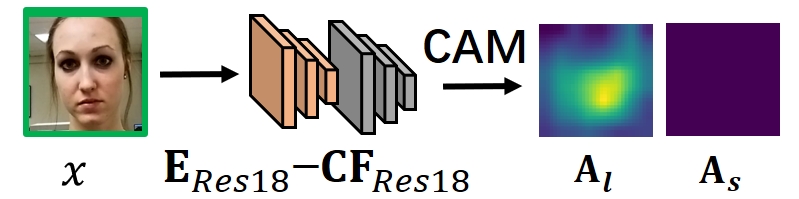}} &
    {\includegraphics[width=5.5cm]{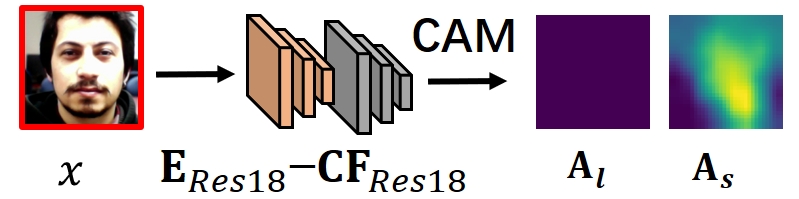}} \\
    (a)&(b) 
    \end{tabular}
\caption{ 
Generation of quasi-ground truth for the two attention estimators \textbf{LE} and \textbf{SE} from (a) a live image and (b) a spoof image.
}
 \label{fig:attention_1}  
\end{figure}

 \paragraph{\textcolor{red}{\textbf{Activation Map Generation}}}
We use Figures \ref{fig:overview} and \ref{fig:attention_1} to explain the proposed idea and the training stage of $\textbf{E}_{Res18}-\textbf{CF}_{Res18}$, \textbf{LE} and \textbf{SE}. 
First, we jointly train the encoder $\textbf{E}_{Res18}$ and the live/spoof classifier $\textbf{CF}_{Res18}$ of Res18 \cite{he2016deep} with LDCformer using 
the liveness loss $\mathcal{L}_{ls}$ in Equation \eqref{eq:liveness_loss}.
Next, we use $\textbf{E}_{Res18}$ and $\textbf{CF}_{Res18}$ to generate the live activation map $\mathbf{A}_l$ and the spoof activation map $\mathbf{A}_s$ of an input image $x$ by, 
 \begin{eqnarray} 
\begin{aligned}
\mathbf{A}_l
&= \textbf{Grad-CAM} (\textbf{CF}_{Res18}( \textbf{E}_{Res18}(x)); y=1), \text{and}\\
\mathbf{A}_s
&= \textbf{Grad-CAM} (\textbf{CF}_{Res18}( \textbf{E}_{Res18}(x)); y=0),
\end{aligned}
\label{eq:produce_activation}  
\end{eqnarray} 

\noindent 
where \textbf{Grad-CAM} indicates the class activation operation. 
In Figure \ref{fig:attention_1}, we see that: (1) the live activation maps $\mathbf{A}_l$ of live images concentrate mostly on facial regions, (2) the spoof activation maps $\mathbf{A}_s$ of live images induce nearly no activation responses, and vice versa for the spoof images. Thus, the activation maps $\mathbf{A}_l$ and $\mathbf{A}_s$ not only preserve the binary live/spoof labels but also offer live/spoof-specific attentions. 

 \paragraph{\textcolor{red}{\textbf{Dual Attention Loss}}}
Next, we use the activation maps $\mathbf{A}_l$ and $\mathbf{A}_s$ as the quasi-ground truth to train the two attention estimators \textbf{LE} and \textbf{SE}. To further constrain the fine-grained feature learning, we use the live activation maps $\mathbf{A}_l$ of live images and the spoof activation maps $\mathbf{A}_s$ of spoof images as their ground-truth, but set all the values of $\mathbf{A}_l$ of spoof images and all the values of $\mathbf{A}_s$ of live images into zeros to strictly constrain the training of \textbf{LE} and \textbf{SE}. 

Finally, by referring to $\mathbf{A}_l$ and $\mathbf{A}_s$, we define the dual attention loss $\mathcal{L}_{dual}$ by, 
\begin{eqnarray}
\mathcal{L}_{dual} = \mathcal{L}_{A_l} +\mathcal{L}_{A_s}
 = ||\mathbf{A}_l - \Bar{\mathbf{A}}_l ||_2+||\mathbf{A}_s - \Bar{\mathbf{A}}_s ||_2,
\label{eq:dual_total}
\end{eqnarray}
\noindent
where $\Bar{\mathbf{A}}_l$ and $\Bar{\mathbf{A}}_s$ are the estimated attentions by \textbf{LE} and \textbf{SE}, respectively.

\begin{figure}  
    \centering
    {\includegraphics[height=2.cm]{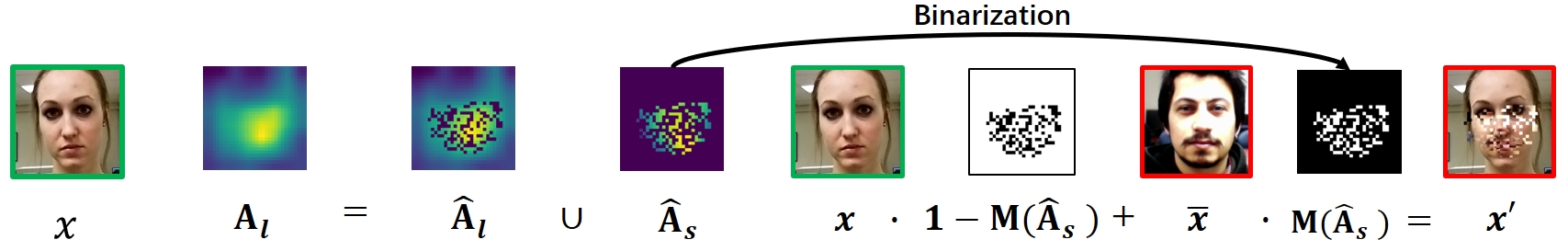}}
\caption{ 
\textcolor{red}{
Generation of the attention-based ground truth ${\mathbf{\hat{A}}}_l$ and ${\mathbf{\hat{A}}}_s$ and the self-challenging images $x^\prime$. This example illustrates the case when $x$ is a live image (i.e. $y=1$) and $\Bar{x}$ is a spoof image. The other case, where $x$ is a spoof image (i.e. $y=0$) and $\Bar{x}$ is a live image, is generated similarly.
} } 
\label{fig:attention_2_3} 
\end{figure}

\subsubsection{Self-Challenging Supervision}
\label{sec:masked_dual_attention} 
In Sec. \ref{sec:Dual-Attention Supervision}, although dual-attention supervision provides a fine-grained quasi-ground truth to guide the model training, the model remains oblivious to any external or previously unseen knowledge beyond the training data \textcolor{red}{(e.g., subtle partial spoofing attacks)}. 
In \cite{huang2020self}, an interesting idea is proposed to challenge the classifier with masked features for improving the feature discriminability. 
Inspired by \cite{huang2020self}, we propose to incorporate a second training strategy by generating self-challenging data to challenge LDCformer  to further improve its feature discriminability. As shown in Figure \ref{fig:overview}, we propose to generate the attention-based ground truth ${\mathbf{\hat{A}}}_l$ and ${\mathbf{\hat{A}}}_s$ and the corresponding self-challenging images $x^\prime$ to supervise LDCformer. 

\begin{figure}  
\centering
    {\includegraphics[height=2cm]{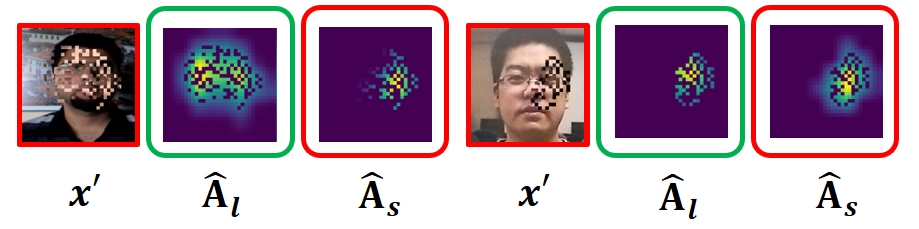}} 
\caption{  
\textcolor{red}{Examples of self-challenging images and their paired live and spoof attentions featuring the mixed-print attacks.
The other case (i.e., the mixed-replay attacks) is generated similarly.
}}
 \label{fig:attention_4}  
\end{figure}

 \paragraph{\textcolor{red}{\textbf{Self-Challenging  Data Generation}}}
We use Figure \ref{fig:attention_2_3} to explain this data generation process from a pair of input images $(x,\Bar{x})$ with different ground-truth labels; that is, either the labels are $y=1$ and $\Bar{y}=0$, or $y=0$ and $\Bar{y}=1$.
First, as shown in Figure \ref{fig:attention_2_3}, after obtaining the live attention $\mathbf{A}_l$ of a live image $x$ (i.e.,  $y=1$), we randomly divide $\mathbf{A}_l$ into two disjoint attentions ${\mathbf{\hat{A}}}_l $ and ${\mathbf{\hat{A}}}_s$. Next, we conduct a binarization operation $\mathbf{M}(\cdot)$ on ${\mathbf{\hat{A}}}_s$ and then use the masked attentions ${\textbf{M}(\mathbf{\hat{A}}}_s)$ to generate the live/spoof-mixed data $x^\prime$ by, 
\begin{eqnarray} 
\begin{aligned}
x^{\prime} & = \mathbbm{1}(y=1)(x \cdot (1-\mathbf{M}(\mathbf{\hat{A}}_s))+\bar{x} \cdot \mathbf{M}(\mathbf{\hat{A}}_s))  \\
& + \mathbbm{1}(y=0)(x \cdot (1-\mathbf{M}(\mathbf{\hat{A}}_l))+\bar{x} \cdot \mathbf{M}(\mathbf{\hat{A}}_l)) 
\end{aligned}
\label{eq:produce_self_challenging_data1}
\end{eqnarray} 

\noindent 
where $\mathbbm{1}(\cdot)$ is the indicator function. Similarly, given a spoof image $x$ (i.e., $y = 0$) and a live image $\bar{x}$, we randomly divide the spoof attention $\mathbf{A}_s$ of $x$ into $\mathbf{A}_s = {\mathbf{\hat{A}}}_l \cup {\mathbf{\hat{A}}}_s $ and use the masked attentions ${\textbf{M}(\mathbf{\hat{A}}}_l)$ to generate their live/spoof-mixed data $x^\prime$ by Equation \eqref{eq:produce_self_challenging_data1}. 
\textcolor{red}{Figure \ref{fig:attention_4}  gives some examples of self-challenging data with mixed-print attacks.}

 \paragraph{\textcolor{red}{\textbf{Self-Challenging Loss}}}
Note that, all these self-challenging data are considered as spoof data in training LDCformer.  
Finally, by referring to ${\mathbf{\hat{A}}}_l$ and ${\mathbf{\hat{A}}}_s$ as the ground truth of live and spoof attentions for the self-challenging data $x^{\prime}$, we define the self-challenging loss $\mathcal{L}_{sc}$ as,
\begin{eqnarray}  
\mathcal{L}_{sc} =\mathcal{L}_{\hat{A}_l} +\mathcal{L}_{\hat{A}_s} =||\mathbf{\hat{A}}_l - \Bar{\mathbf{A}}^\prime_l ||_2+||\mathbf{\hat{A}}_s - \bar{\mathbf{A}}^\prime_s ||_2,
\label{eq:ls_masked_attention_loss}
\end{eqnarray} 
\noindent
where $\Bar{\mathbf{A}}^\prime_l$ and $\Bar{\mathbf{A}}^\prime_s$ are the estimated attentions of the self-challenging image $x^{\prime}$ by  \textbf{LE} and \textbf{SE}, respectively.

\begin{figure} 
    \begin{minipage}[b]{1.0\linewidth}
      \centering{ 
      \includegraphics[width=12cm]{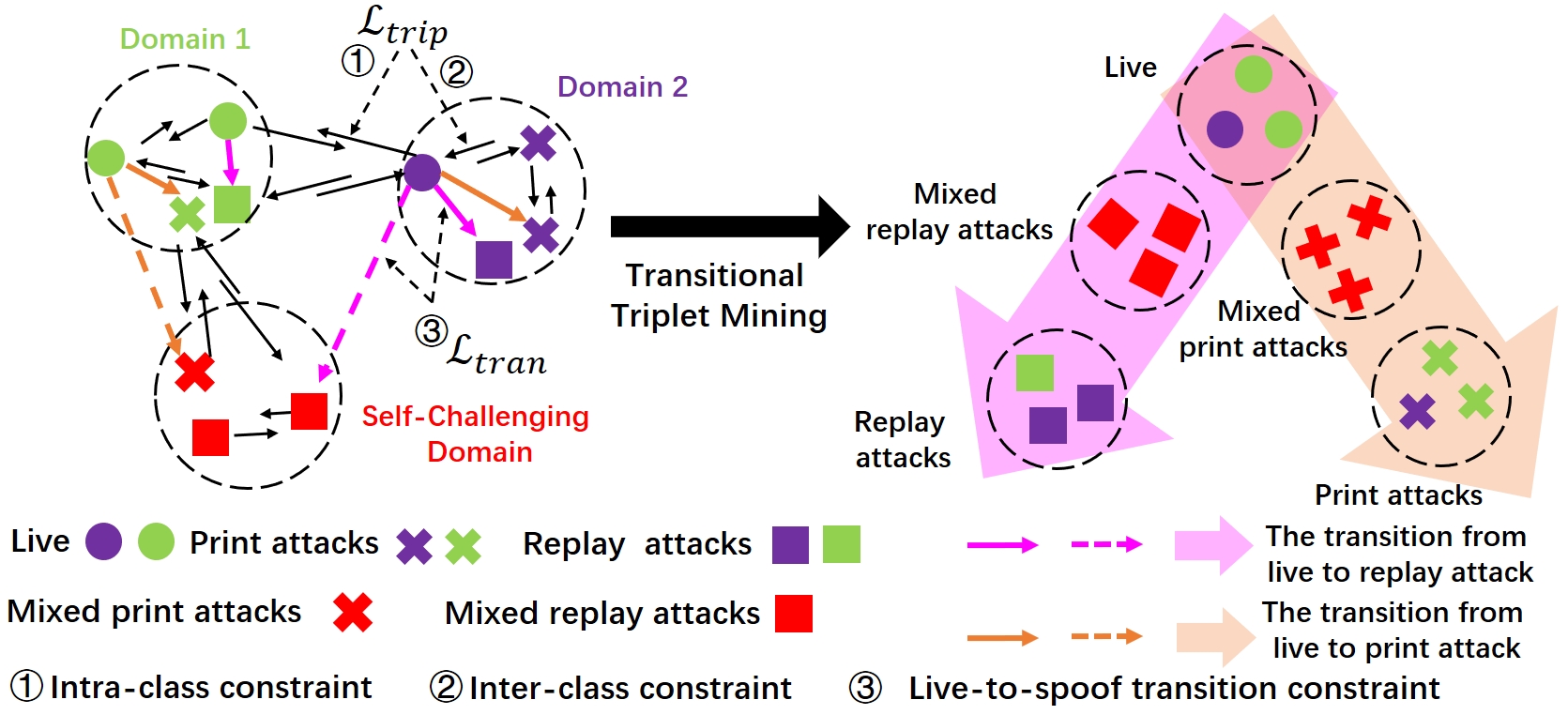} 
      }
    \end{minipage}   
\caption{
Illustration of the proposed transitional triplet mining.   
}
 \label{fig:triplet_mining} 
\end{figure}

\subsubsection{Transitional Triplet Mining}
\label{sec:Textural Triplet Mining} 
The idea of triplet mining has been adopted in our previous work \cite{huang2022learnable} to encourage models on learning domain-invariant representation. 
The basic idea is to separate samples of different classes while aggregating samples of the same class from different domains.
While triplet mining adopted in \cite{huang2022learnable} effectively enabled the model to narrow the domain gap, it involves only the training domains in the learning process but is completely unaware of  previously unseen domain knowledge. 
Therefore, in this paper, we further include the self-challenging data in Section \ref{sec:masked_dual_attention} into the triplet mining to widen the domain scope. 
Moreover, we propose a novel transitional consistency loss by examining the relationship between different spoof attacks and live images to enforce the model to learn domain-invariant and domain-generalized features.

\paragraph{\textcolor{red}{\textbf{Transitional Triplet Loss}}}
We assume each benchmark dataset indicates one domain and also consider the self-challenging data as a new domain. 
In addition, because images under different presentation attacks have different characteristics, we regard the spoof images under different attacks as different classes. 
Let $X$ denote the set of training images from different domains, and $X^\prime$ denote the generated self-challenging data.
In this subsection, we label all the images in $X$ and the self-challenging images in $X^\prime$ into five classes, i.e., live, print attack, replay attack, mixed-print attack, and mixed-replay attack, as shown in Figure  \ref{fig:triplet_mining}.
We include both the triplet loss $\mathcal{L}_{trip}$ and the transitional consistency loss $\mathcal{L}_{tran}$ to define the transitional triplet loss $\mathcal{L}_{tran-trip}$ by,
\begin{eqnarray}  
\mathcal{L}_{tran-trip} =\mathcal{L}_{trip} +\mathcal{L}_{tran}.
\label{eq:transitional triplet mining}
\end{eqnarray} 
\noindent 
 
\paragraph{\textcolor{red}{\textbf{Triplet Loss}}} First, we define the triplet loss $\mathcal{L}_{trip}$. As illustrated in Figure \ref{fig:triplet_mining}, our goal is to separate inter-class pairs from intra-class pairs by least a margin $\alpha$ to learn domain-invariant features from the training set $x$ and the self-challenging dataset $X^\prime$ by,
\begin{eqnarray}  
\mathcal{L}_{trip}= 
\sum_{\forall \, x_a \in (X \cup X^\prime)}  \Bigl (
 \| \textbf{z}^{N_1+N_2}_{a}  - \textbf{z}^{N_1+N_2}_{p} \|_2  
+  (\alpha -  \| \textbf{z}^{N_1+N_2}_{a}  - \textbf{z}^{N_1+N_2}_{n}\|_2 ) \Bigl ), 
\label{eq:triplet_loss}  
\end{eqnarray} 

\noindent 
where $\textbf{z}^{N_1+N_2}_{a}$ is the LDCformer feature of an anchor image $x_a$,  $\textbf{z}^{N_1+N_2}_{p}$ and $\textbf{z}^{N_1+N_2}_{n}$ are the features of a positive image $x_p$ (i.e., with the same class label as $x_a$) and a negative image $x_n$ (i.e., with the class label different from $x_a$), respectively. In Equation \eqref{eq:triplet_loss}, we define the margin $\alpha$ by, 
\begin{eqnarray}
\alpha=\left\{
\begin{aligned}
0.05, \, & if \, x_a \in X^\prime \ or \ x_n \in X^\prime\\
0.1, \, & \, otherwise;
\end{aligned}
\right.
\label{eq:s_h}
\end{eqnarray}

\noindent 
Note that, we set different margins for the original data $X$ and the self-challenging data $X^\prime$ when calculating the triplet loss in Equation \eqref{eq:triplet_loss}, because $X^\prime$ are generated by mixing both live and spoof images and thus possesses characteristics from both live and spoof images. 
  
\paragraph{\textcolor{red}{\textbf{Transitional Consistency Loss}}} Next, we define the transitional consistency loss $\mathcal{L}_{tran}$. 
Note that, we consider the self-challenging data $X^\prime$ as a new domain, because $X^\prime$ possess its own distinctive domain characteristics. 
However, since we generate $X^\prime$ by mixing live and spoof images, the generated images $x^\prime$ usually exhibit characteristics from both the live and spoof images in $X$.  
Moreover, the images $x^\prime$ generated by mixing a live image and a spoof image of print attack should exhibit similar spoof characteristics of print attack, and images $x^\prime$ generated from mixing a live image and a spoof image of replay attack should also exhibit the spoof characteristics of replay attack.
Therefore, to further enforce the model to learn live/spoof discriminative features, we impose a consistent constraint between the live-to-spoof transition and the live-to-mixed-spoof transition. 
As illustrated in Figure \ref{fig:triplet_mining}, we define $\mathcal{L}_{tran}$ to maximize the cosine similarity between the live-to-replay-attack transition and the live-to-mixed-replay-attack transition as well as the cosine similarity between the live-to-print-attack and the live-to-mixed-print-attack for learning domain-generalized features by, 
\begin{eqnarray}
\mathcal{L}_{tran} &= & \sum_{\forall \, x_l \in  X_l }  \bigg( \sum_{\substack{\forall x_s \in  X_p , \\ \forall x_s^\prime \in  X^\prime_p }}  (1 -  {sim}(\mathbf{z}^{N1+N2}_{s} - \mathbf{z}^{N1+N2}_{l} , \mathbf{z}^{N1+N2}_{s^\prime } - \mathbf{z}^{N1+N2}_{l}) ) + 
\nonumber \\ 
&&   \sum_{\substack{\forall x_s \in  X_r , \\ \forall x_s^\prime \in  X^\prime_r }}  (1 -  {sim}(\mathbf{z}^{N1+N2}_{s} - \mathbf{z}^{N1+N2}_{l} , \mathbf{z}^{N1+N2}_{s^\prime } - \mathbf{z}^{N1+N2}_{l})) \bigg ) 
\label{eq:transitional_loss}  
\end{eqnarray}  

\noindent
where $\mathbf{z}^{ N1+N2}_{l}$ is the feature of a live image $x_l$,  $\mathbf{z}^{ N1+N2}_{s}$ and $\mathbf{z}^{ N1+N2}_{s^\prime}$ denote the feature of a spoof image $x_s$ of certain attack and a mixed-attack $x^{\prime}_s$, respectively. 

To sum up, by minimizing the proposed transitional triplet loss $\mathcal{L}_{tran-trip}$, 
we encourage LDCformer: 1) to learn domain-invariant features by narrowing the distance between different domains via the triplet loss $\mathcal{L}_{trip}$, and  
2) to learn domain-generalized features by maintaining the consistency between live-to-spoof and live-to-mixed-attack transitions via the transitional consistency loss $\mathcal{L}_{tran}$.  

\subsection{Training and Testing} 
\label{sec:Train and Testing} 
In the training stage, we include the liveness loss $\mathcal{L}_{ls}$, the dual attention loss $\mathcal{L}_{dual}$, the self-challenging loss $\mathcal{L}_{sc}$, and the transitional triplet loss $\mathcal{L}_{tran-trip}$ into the total loss $\mathcal{L}_{T}$ to collaboratively train LDCformer, the auxiliary network $\textbf{E}_{Res18}-\textbf{CF}_{Res18}$, and the two attention estimators \textbf{LE} and \textbf{SE}. The total loss is thus defined by, 
\begin{eqnarray}
\mathcal{L}_{T} = \mathcal{L}_{ls} + \beta  \mathcal{L}_{dual} + \gamma  \mathcal{L}_{sc}
\label{eq:loss_total} + \delta \mathcal{L}_{tran-trip} 
\end{eqnarray}
where $\beta$, $\gamma$, and  $\delta$ are the weight factors. In all our experiments, we empirically set $\beta=0.004$, $\gamma=0.004$, and $\delta=0.1$. 
 
In the inference stage, we use only LDCformer to conduct live/spoof classification.
We follow previous face anti-spoofing methods \cite{liu2018learning,huang2022face,shao2020regularized} to measure the detection score in terms of the output from the classifier \textbf{CF} in LDCformer.  
For each test image $x$, we define the detection score $s $ by,  
\begin{eqnarray}  
s =  \mathbf{LDCformer}(x).
\label{eq:prediction_score}
\end{eqnarray}

\begin{table}[t]
\color{red}
\normalsize  
\large 
\caption{\textcolor{red}{Summary of different metrics used in different scenarios.}}
\label{tab:metric}
\resizebox{\columnwidth}{!}{ 
\begin{tabular}{|c|l|c|}
\hline
Metrics & \multicolumn{1}{c|}{Description} & Testing scenarios\\ \hline
APCER (\%) $\downarrow$ & \begin{tabular}{l}The Attack Presentation Classification Error Rate (APCER) measures the error rate of \\ incorrectly classifying spoof attempts as genuine (live) presentations. \end{tabular} & Intra-domain testing   \\ \hline
BPCER (\%) $\downarrow$  & \begin{tabular}{l} The Bona Fide Presentation Classification Error Rate (BPCER) measures the error rate \\ of incorrectly classifying genuine (live) presentations as spoof attempts. \end{tabular}  &  Intra-domain testing   \\ \hline
\rule[-1ex]{0pt}{3.5ex} ACER  (\%) $\downarrow$ & \begin{tabular}{l} The Average Classification Error Rate (ACER) is the average of APCER and  BPCER. \end{tabular}  & Intra-domain testing  \\ \hline
HTER  (\%) $\downarrow$  & \begin{tabular}{l} The practical significance of the Half Total Error Rate (HTER) is equivalent to that of \\ the Average Classification Error Rate (ACER). \end{tabular}  &   Cross-domain testing \\ \hline
AUC   (\%) $\uparrow$  & \begin{tabular}{l} The Area Under the Curve (AUC) is used to evaluate the model's ability to distinguish \\between genuine users and spoof attacks across a range of thresholds. \end{tabular}   & Cross-domain testing\\ \hline
\end{tabular}
}
\end{table}

\section{Experiments}
\subsection{Experimental Setting}
\subsubsection{Datasets and Evaluation Metrics}

We conduct extensive experiments on the following face anti-spoofing databases: 
\textbf{OULU-NPU} \cite{boulkenafet2017oulu} (denoted by O), \textbf{MSU-MFSD} \cite{wen2015face} (denoted by M), \textbf{CASIA-MFSD} \cite{zhang2012face} (denoted by C), \textbf{Idiap Replay-Attack} \cite{chingovska2012effectiveness} (denoted by I), \textbf{SiW} \cite{liu2018learning}, \textbf{3DMAD} \cite{erdogmus2014spoofing}, \textbf{HKBU-MARs} \cite{liu20163d}, \textbf{CASIA-3DMask} \cite{yu2020fas} and \textbf{PADISI-Face} \cite{rostami2021detection}. 
\textcolor{red}{The batch size is set to 18 in all experiments.}
To have a fair comparison with previous methods, we follow the original protocols to conduct intra- and cross-testings on these datasets and \textcolor{red}{report the results using the same evaluation metrics, including APCER (\%)   $\downarrow$,  BPCER (\%)   $\downarrow$,  ACER (\%),  HTER (\%)   $\downarrow$, and AUC (\%) $\uparrow$, as summarized in Table~\ref{tab:metric}.}

\begin{table}[t]
\centering
\caption{Ablation study on \textbf{{[}I,C,M{]} $\rightarrow$ O}, using different numbers of $\textbf{E}_{LDCformer}$ in LDCformer. 
The evaluation metrics are HTER(\%) $\downarrow$ and AUC(\%) $\uparrow$.\label{tab:ablation_study_ldc_layer}
}
\resizebox{\columnwidth}{!}{%
\begin{tabular}{ccccccccccccccc}
\hline
\textbf{Method} &   $N_1$         				& 0 & 1 & 2 & 3 & 4 & 5 & 6 & 7 & 8 & 9 & 10 & 11 & 12 \\ \hline
\multirow{2}{*}{LDCformer} 	& HTER & 15.67  & 12.70  & \textbf{12.21}  & 12.23   & 12.99  & 12.84  & 12.55  & 12.84  & 12.83  & 12.78  & 12.40   & 12.85   & 13.96   \\ \cline{2-15} 
                               					& AUC  & 88.71  & 93.04  & \textbf{94.36}  & 93.67  & 92.68  & 93.89  & 94.11  & 94.30  & 94.15  & 94.35  & 94.17   & 94.21   & 92.79   \\ \hline
\end{tabular}%
}
\end{table}

\begin{table}[t]
\centering 
\scriptsize
\caption{Ablation study on \textbf{{[}I,C,M{]}$\rightarrow$ O} using different convolutions to cooperate with ViT for FAS. 
The evaluation metrics are HTER(\%) $\downarrow$ and AUC(\%) $\uparrow$.
}
\label{tab:ablation study different convolution}
\resizebox{\columnwidth}{!}{%
\begin{tabular}{ccccccccc}
\hline
\multicolumn{2}{c}{\multirow{2}{*}{\textbf{Method}}}         &  \multirow{2}{*}{ViT\cite{dosovitskiy2021an}}  & \multicolumn{1}{c}{ViT +} & \multicolumn{1}{c}{ViT +} & \multicolumn{1}{c}{ViT +} & \multicolumn{1}{c}{ViT +} & \multicolumn{1}{c}{ViT +} & \multicolumn{1}{c}{ViT +}  \\ 
\multicolumn{2}{c}{}                                &     & CNN                       & LBC \cite{juefei2017local}                       & Sobel \cite{wang2020deep}                     & CDC \cite{yu2020searching}                       & C-CDC \cite{yu2021dual}                     & LDC \cite{huang2022learnable}                                    \\ \hline
\multicolumn{1}{c}{}                         & HTER   & 15.67 & 14.12 & 13.67 & 13.48 & 13.30 & 12.99 & \textbf{12.21}                           \\ \cline{2-9} 
\multicolumn{1}{c}{}                         & AUC  & 88.71 & 89.59 & 90.45 & 91.54 & 90.93 & 90.92 & \textbf{94.36}                           \\ \hline
\end{tabular}%
}
\end{table}

\begin{table}[t]
\centering
\color{red}
\scriptsize
\setlength\tabcolsep{4pt}
\caption{Ablation study on the cross-domain protocol \textbf{{[}I,C,M{]} $\rightarrow$ O} and the intra-domain protocol  \textbf{Oulu} \textbf{Prot.} 4, using different combinations of loss terms.
The evaluation metrics are HTER(\%) $\downarrow$ and AUC(\%) $\uparrow$.
\label{tab:ablation_study_loss}}
\begin{tabular}{ccccccccc}
\hline
\multicolumn{6}{c}{Total loss $\mathcal{L}_T$} & \multicolumn{2}{c}{\textbf{{[}I,C,M{]} $\rightarrow$ O}} & \textbf{Oulu} \textbf{Prot.} 4 \\ \hline

\multicolumn{1}{c}{$\mathcal{L}_{ls}$} & \multicolumn{1}{c}{$\mathcal{L}^\prime_{dual}$} & \multicolumn{1}{c}{$\mathcal{L}_{dual}$} & \multicolumn{1}{c}{$\mathcal{L}_{sc}$} & $\mathcal{L}_{trip}$  
& $\mathcal{L}_{tran-trip}$ &  \multicolumn{1}{c}{HTER (\%) $\downarrow$} & AUC (\%) $\uparrow$ &  ACER (\%) $\downarrow$  \\ \hline

\multicolumn{1}{c}{\checkmark} & \multicolumn{1}{c}{} & \multicolumn{1}{c}{} & \multicolumn{1}{c}{} & \multicolumn{1}{c}{} &  & \multicolumn{1}{c}{12.21} & 94.36 & 2.51 $\pm$ 0.66   \\ \hline

\multicolumn{1}{c}{\checkmark} & \multicolumn{1}{c}{\checkmark} & \multicolumn{1}{c}{} & \multicolumn{1}{c}{} & \multicolumn{1}{c}{} &  & \multicolumn{1}{c}{11.29} & 94.79 & 2.03 $\pm$ 0.11  \\ \hline

\multicolumn{1}{c}{\checkmark} & \multicolumn{1}{c}{} & \multicolumn{1}{c}{\checkmark} & \multicolumn{1}{c}{} & \multicolumn{1}{c}{} &  & \multicolumn{1}{c}{10.20} & 96.14 & 1.69 $\pm$ 0.68 \\ \hline

\multicolumn{1}{c}{\checkmark} & \multicolumn{1}{c}{} & \multicolumn{1}{c}{} & \multicolumn{1}{c}{\checkmark} & \multicolumn{1}{c}{} &  & \multicolumn{1}{c}{9.66} & 96.57 & 1.43 $\pm$ 0.51  \\ \hline

\multicolumn{1}{c}{\checkmark} & \multicolumn{1}{c}{} & \multicolumn{1}{c}{} & \multicolumn{1}{c}{} & \multicolumn{1}{c}{} & \checkmark & \multicolumn{1}{c}{9.43} & 96.60 & 1.47 $\pm$ 1.97  \\ \hline

\multicolumn{1}{c}{\checkmark} & \multicolumn{1}{c}{} & \multicolumn{1}{c}{\checkmark} & \multicolumn{1}{c}{\checkmark} & \multicolumn{1}{c}{} &  & \multicolumn{1}{c}{9.02} & 96.92 & 1.19 $\pm$ 1.18 \\ \hline

\multicolumn{1}{c}{\checkmark} & \multicolumn{1}{c}{} & \multicolumn{1}{c}{\checkmark} & \multicolumn{1}{c}{\checkmark} & \multicolumn{1}{c}{\checkmark} &  & \multicolumn{1}{c}{8.59} & 96.96 & 0.69 $\pm$ 0.19 \\ \hline

\multicolumn{1}{c}{\checkmark} & \multicolumn{1}{c}{} & \multicolumn{1}{c}{\checkmark} & \multicolumn{1}{c}{\checkmark} & \multicolumn{1}{c}{} & \checkmark & \multicolumn{1}{c}{\textbf{6.77}} & \textbf{97.89} & 
\textbf{0.25 $\pm$ 0.42} \\ \hline
\end{tabular}
\end{table}

\begin{table}[t]
\centering
\scriptsize 
\setlength{\tabcolsep}{2pt}
\caption{Comparison of intra-domain face presentation attack detection on \textbf{OULU-NPU}.
The evaluation metrics are APCER(\%) $\downarrow$, BPCER(\%) $\downarrow$, and ACER(\%) $\downarrow$.
$\dag$ indicates the transformer-based method. 
\label{tab:oulu_intra}
}
\resizebox{\columnwidth}{!}{
\begin{tabular}{ccccc cccc}
\hline
\textbf{Method} & \textbf{Prot.} & APCER & BPCER & ACER & \textbf{Prot.} &  APCER & BPCER & ACER  \\ \hline




\multicolumn{1}{c}{RAEDFL \cite{huang2022face} (\textit{ACPR 21})} & \multirow{9}{*}{1} & 1.67 & 0.00 & 0.83 						& \multirow{9}{*}{3} &  1.38$\pm$1.78 & 0.28$\pm$0.68 & 0.83$\pm$0.86 \\



ViTranZFAS \cite{george2021effectiveness} $\dag$ (\textit{IJCB 21}) & & 0.0 & 0.8 & {0.4} 						& &  1.84$\pm$1.73 & 1.50$\pm$1.67 & 1.66$\pm$1.53\\

PatchNet \cite{wang2022patchnet} (\textit{CVPR 22}) & & {0.0} & {0.0} & \textbf{0.0}							& &  1.8$\pm$1.47 & 0.56$\pm$1.24 & 1.18$\pm$1.26 \\

LDCN \cite{huang2022learnable} (\textit{BMVC 22}) & & 0.0 & 0.0 & \textbf{0.0} 									& &  4.55$\pm$4.55 & 0.58$\pm$0.91 & 2.57$\pm$2.67 \\ 

TransFAS \cite{wang2022face} $\dag$ (\textit{TBBIS 22}) & & 0.8 & 0.0 & 0.4 									& &  0.6$\pm$0.7 & 1.1$\pm$2.5 & 0.9$\pm$1.1 \\

TTN-S \cite{wang2022learning} $\dag$ (\textit{TIFS 22}) & & 0.4 & 0.0 & 0.2 									& &  1.0$\pm$1.1 & 0.8$\pm$1.3 & 0.9$\pm$0.7 \\  

CSM-GAN \cite{wu2022covered} (\textit{PR 22}) & & 0.14 & 0.56 & 0.35 									& &  0.50$\pm$0.97 & 2.83$\pm$1.38 & 1.67$\pm$1.05 \\  

FDGAL \cite{huang2023face} (\textit{PR 23}) & & 0.9 & 1.0 & 1.0									& &  2.6$\pm$1.3 & 0.3$\pm$0.6 & 1.4$\pm$0.5 \\ 

\textcolor{red}{LGON \cite{wang2023learnable} (\textit{PR 23})} & & 1.5 & 0.0 & 0.8									& &  1.3$\pm$0.9 & 0.8$\pm$0.9 & 1.0$\pm$0.6 \\ 


  \cline{1-9}
\multicolumn{1}{c}{LDCformer $\dag$} & &  0.00 & \multicolumn{1}{c}{0.00} & \multicolumn{1}{c}{\textbf{0.00}} & & \multicolumn{1}{c}{0.75$\pm$0.94} & \multicolumn{1}{l}{0.06$\pm$0.13} & \multicolumn{1}{l}{\textbf{0.21$\pm$0.43}}\\		


 \hline






	RAEDFL \cite{huang2022face} (\textit{ACPR 21})  & \multirow{9}{*}{2} & 0.69 & 1.67 & 1.18 							& \multirow{9}{*}{4} &	 	 5.41$\pm$6.40 & 2.50$\pm$2.74 & {3.96$\pm$3.90} \\



  	ViTranZFAS \cite{george2021effectiveness} $\dag$ (\textit{IJCB 21}) & & 1.3 & 1.9 & 1.6 	& &	 	 2.67$\pm$1.63 & 3.50$\pm$3.51& 3.08$\pm$2.21\\

 	PatchNet \cite{wang2022patchnet} (\textit{CVPR 22})&  & 0.8 & 1.0 & 0.9 						& & 	 2.5$\pm$3.81 & 3.33$\pm$3.73 & 2.90$\pm$3.00\\

    LDCN \cite{huang2022learnable} (\textit{BMVC 22}) & & 0.8 & 1.0 & 0.9 						& & 	 4.50$\pm$1.48 & 3.17$\pm$3.49 & 3.83$\pm$2.12 \\

 	TransFAS \cite{wang2022face} $\dag$ (\textit{TBBIS 22}) & & 1.5 & 0.5 & 1.0 					& & 	 2.1$\pm$2.2 & 3.8$\pm$3.5 & 2.9$\pm$2.4 \\

 	TTN-S \cite{wang2022learning} $\dag$ (\textit{TIFS 22}) & & 0.4 & 0.8 & 0.6 					& & 	 3.3$\pm$2.8 & 2.5$\pm$2.0 & 2.9$\pm$1.4 \\  


        CSM-GAN \cite{wu2022covered} (\textit{PR 22}) & & 0.69 & 1.67 & 1.18									& &  2.22$\pm$1.77 & 8.29$\pm$4.18 & 5.26$\pm$2.88 \\ 
        
        FDGAL \cite{huang2023face} (\textit{PR 23}) & & 1.0 & 1.1 & 1.1									& &  3.8$\pm$5.2 & 5.8$\pm$4.9 & 4.8$\pm$4.3 \\ 
    \textcolor{red}{LGON \cite{wang2023learnable} (\textit{PR 23})} & & 2.3 & 1.4 & 1.9									& &  3.3$\pm$2.6& 3.3$\pm$4.1 & 3.3$\pm$2.6 \\ 
        
  \cline{1-9}
	LDCformer $\dag$ & & 0.36 & 0.28 & \textbf{0.32} 											& & 	 0.38$\pm$0.63 &0.13$\pm$0.21 & \textbf{0.25$\pm$0.42} \\
 \hline
\end{tabular}
}
\end{table} 

\subsubsection{Network Architecture and Implementation Details} 
We implement  LDCformer  by Pytorch and set $N_1=2$ (i.e., the number of LDCformer encoders) and $N_2=10$ (i.e., the number of standard Transformer encoders). We build the two attention estimators \textbf{LE} and \textbf{SE} using three convolutional blocks, where each block consists of a convolutional layer, a batch normalization layer, and a ReLU activation function. We set a constant learning rate of 5$\mathbf{e-}$4 with Adam optimizer up to 100 epochs to train \textbf{MLP} and \textbf{E}$_{Res34}$, and a constant learning rate of 1$\mathbf{e-}$4 with Adam optimizer up to 200 epochs to train LDCformer, \textbf{LE}, and \textbf{SE}. 
To enable direct comparisons with previous methods under the same ViT backbone, 
we follow \cite{george2021effectiveness} to adopt the standard ViT-Base backbone to evaluate the proposed method.  
\textcolor{red}{
The codes are available at  \url{https://github.com/Pei-KaiHuang/LDCformer_Ext}.
}

\subsection{Ablation Study}

\subsubsection{Using Different Numbers of LDCformer Encoder $\textbf{E}_{LDCformer}$}  

In Table \ref{tab:ablation_study_ldc_layer}, we evaluate how $N_1$ affects the performance of LDCformer.
Because the standard ViT consists of 12 standard Transformer encoders $\textbf{E}_{ViT}$, we set the range of $N_1$ for $\textbf{E}_{LDCformer}$ from 0 to 12. 
Here we train LDCformer simply using the liveness loss $\mathcal{L}_{ls}$ on the three \textbf{{[}I,C,M{]}} datasets and test LDCformer  on the other one \textbf{O}. 
Note that, this cross-domain protocol \textbf{{[}I,C,M{]}$\rightarrow$ O} is very challenging because the three datasets \textbf{{[}I,C,M{]}} are much smaller than the dataset \textbf{O}.


As shown in Table \ref{tab:ablation_study_ldc_layer}, first, we notice that all the settings of LDCformer (i.e., $N_1=1-12$) outperform the original ViT (i.e., $N_1=0$).
This performance improvement shows that incorporating LDC into ViT indeed leverages its long-range modeling capacity with local and discriminative features.
Next, among all the other settings of $N_1>0$, using $N_1=2$ in LDCformer achieves the best performance with the lowest ACER and the highest AUC. 
We suspect that, when including only one $\textbf{E}_{LDCformer}$ ( i.e., $N_1=1$) in LDCformer, there involve not many low-level features into ViT. On the other hand, when including more than two encoders $\textbf{E}_{LDCformer}$, 
the deeper layers of neural network can no longer preserve low-level features; therefore, if adding more LDCformer encoders $\textbf{E}_{LDCformer}$, the LDCformer encoders $\textbf{E}_{LDCformer}$ in deeper layers of LDCformer extract no further low-level features with distinguishing characteristics.  Note that, this result also coincides with the hypothesis mentioned in  \cite{xiao2021early} that the existence of convolutional layer in lower layers helps to improve the performance on ViT.
As a result, LDCformer does not improve in performance by including more $\textbf{E}_{LDCformer}$.
Therefore, from this ablation study, we empirically set $N_1=2$ in LDCformer for all the following experiments.


\subsubsection{Incorporating Different Convolutional Operations into ViT} 
In Table \ref{tab:ablation study different convolution}, we evaluate the effectiveness of incorporating different convolutional operations in ViT. In this experiment, we replace LDC\cite{huang2022learnable} in $\textbf{E}_{LDCformer}$ with vanilla convolution, Local Binary Convolution \cite{juefei2017local}, Sobel Convolution \cite{wang2020deep},  Central Difference Convolution (CDC) \cite{yu2020fas,yu2020searching},  and Dual-Cross Central Difference Convolution (C-CDC) \cite{yu2021dual}.
These models are trained using the liveness loss $\mathcal{L}_{ls}$ on the three \textbf{{[}I,C,M{]}} datasets and then are tested on the other one \textbf{O}.

From Table \ref{tab:ablation study different convolution}, the improved performance of ViT + CNN over ViT shows that combining local convolutional features indeed contributes to better representation capability of ViT. 
The other cases, including ViT + LBC / Sobel / CDC / C-CDC / LDC, all consistently improve ViT and outperform ViT + CNN. 
These results demonstrate that, using either pre-defined local descriptors, i.e., LBC, Sobel, CDC, and C-CDC, or learnable local descriptors, i.e., LDC, 
enhance discriminability of ViT + CNN on learning characteristic features for FAS.
In particular,  LDCformer achieves the best performance; and this outperformance verifies that LDC \cite{huang2022learnable} better adapts to learn various textural details than the pre-defined descriptors.
  
\subsubsection{Evaluation of Different Loss Terms}
\textcolor{red}{
In Table \ref{tab:ablation_study_loss}, we compare using different combinations of loss terms to train LDCformer and evaluate the performance on the challenging cross-domain protocol \textbf{{[}I,C,M{]} $\rightarrow$ O} and the intra-domain protocol \textbf{Oulu} \textbf{Prot.} 4.
}
Here, we call the model trained using only the liveness loss $\mathcal{L}_{ls}$ as the LDCformer baseline. 
In addition, we also compare replacing the proposed dual attention loss $\mathcal{L}_{dual}$ (in Sec.\ref{sec:Dual-Attention Supervision} ) with the pre-trained dual attention supervision (denoted by $\mathcal{L}^\prime_{dual}$)  proposed in our previous work \cite{huang2022learnable}.

\begin{figure}  
    \centering
    \begin{tabular}{ccc} 
    {\includegraphics[height=4cm]{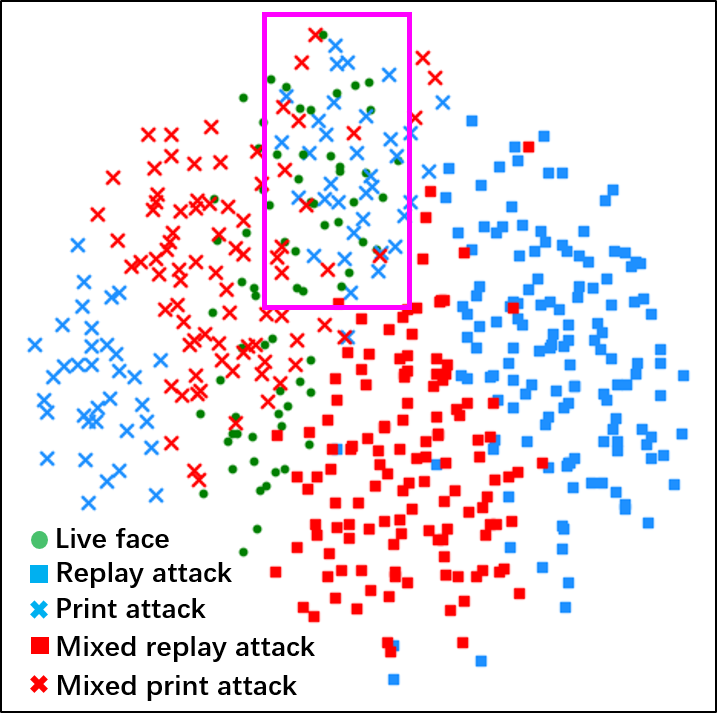}} &
    {\includegraphics[height=4cm]{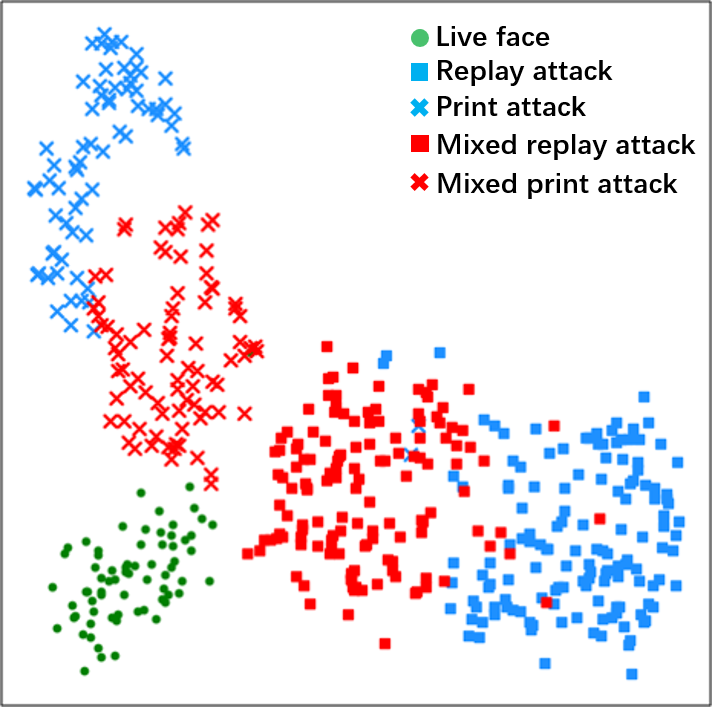}} \\
    (a)&(b)
    \end{tabular} 
\caption{  $t$-SNE visualization of latent features on the challenging cross-domain protocol \textbf{{[}I,C,M{]} $\rightarrow$ O}, using (a) the triplet mining alone and (b) the proposed transitional triplet mining.
The misclassified samples are highlighted by the purple box. 
} 
 \label{fig:tsne}  
\end{figure}

First, inclusion of $\mathcal{L}_{dual}$ with $\mathcal{L}_{ls}$ not only outperforms the baseline but also much improves the performance over the case of $\mathcal{L}_{ls}+\mathcal{L}^\prime_{dual}$. This result verifies that, even without resorting to the pre-trained auxiliary network \cite{huang2022learnable}, the proposed dual-attention supervision effectively improves LDCformer on learning fine-grained features.
Next, when including $\mathcal{L}_{dual} + \mathcal{L}_{sc}$ with the baseline LDCformer, we show that self-challenging supervision, by generating challenging spoof data, further improves the performance.
When additionally including $\mathcal{L}_{trip}$, 
we show that triplet mining is indeed effective on enforcing LDCformer to narrow the domain gap.
Finally, when further including $\mathcal{L}_{tran-trip}$, we achieve the best performance.
In Figures \ref{fig:tsne} (a) and (b), we use $t$-SNE to visualize the live/spoof features learned using only the triplet loss $\mathcal{L}_{trip}$ and the proposed transitional triplet loss $\mathcal{L}_{tran-trip}$, respectively.
From Figure \ref{fig:tsne} (b), we see much enhanced domain-generalization capability of the latent features learned using $\mathcal{L}_{tran-trip}$ than using the triplet loss $\mathcal{L}_{trip}$ alone  (Figure \ref{fig:tsne} (a)) on the challenging cross-domain protocol \textbf{{[}I,C,M{]} $\rightarrow$ O}. 

\textcolor{red}{In addition, by comparing the cases of $\mathcal{L}_{ls} + \mathcal{L}_{dual}$, $\mathcal{L}_{ls} + \mathcal{L}_{sc}$,  and $\mathcal{L}_{ls} + \mathcal{L}_{tran-trip}$, we also investigate the effectiveness of each training strategy individually.
First, we see that self-challenging data ($\mathcal{L}_{ls} + \mathcal{L}_{sc}$) indeed encourages LDCformer to focus on learning discriminative liveness features, resulting in improved performance compared to the case of $\mathcal{L}_{ls} + \mathcal{L}_{dual}$.
Next, by comparing the cases of $\mathcal{L}_{ls} + \mathcal{L}_{dual}$ vs. $\mathcal{L}_{ls} + \mathcal{L}_{tran-trip}$ and  $\mathcal{L}_{ls} + \mathcal{L}_{sc}$ vs. $\mathcal{L}_{ls} + \mathcal{L}_{tran-trip}$, because $  \mathcal{L}_{tran-trip}$ increases the generalization ability of LDCformer, we see that the case $\mathcal{L}_{ls} + \mathcal{L}_{tran-trip}$ significantly improves the performance under  the challenging cross-domain protocol \textbf{{[}I,C,M{]} $\rightarrow$ O}. 
} 

All the results shown in Table \ref{tab:ablation_study_loss} and in Figure  \ref{fig:tsne} verify that the proposed training strategies described in Section \ref{sec:Training Strategies} successfully enable LDCformer to learn domain-invariant and domain-generalized features for FAS.  

\subsection{Experimental Comparisons}

\subsubsection{Intra-Domain Testing} 
In Table \ref{tab:oulu_intra}, we conduct intra-domain testing on \textbf{OULU-NPU} \cite{boulkenafet2017oulu} \textcolor{red}{to evaluate performance in detecting full-face 
 spoof attacks. That is, the spoof attacks in this experiment cover the entire facial regions (i.e., full-face spoof attacks, as shown in Figure~\ref{fig:partial spoof attacks}.)} 
In \textbf{OULU-NPU} \cite{boulkenafet2017oulu}, there are four challenging protocols using different spoof media and capture devices in unseen environments to evaluate FAS models.
As shown in Table \ref{tab:oulu_intra}, LDCformer outperforms previous methods and  achieves state-of-the-art performance on all the protocols. 
In comparison with the recent transformer-based methods  
\cite{george2021effectiveness,wang2022face,wang2022learning}
(denoted by $\dag$ in Tables \ref{tab:oulu_intra}),
LDCformer achieves state-of-the-art performance on four \textbf{OULU-NPU} protocols. 
\textcolor{red}{
In particular, the unsatisfactory performance in \cite{george2021effectiveness} results from constraining a standard ViT with only cross-entropy loss.
In \cite{wang2022face,wang2022learning}, the authors included auxiliary depth information to supervise ViT training in learning fine-grained liveness features; thus,  its performance much improves over \cite{george2021effectiveness}. 
In summary, the superior performance in Table \ref{tab:oulu_intra}
verifies that LDCformer, trained collaboratively with the proposed strategies, indeed benefits from dual-attention and self-challenging supervision to learn discriminative, fine-grained liveness features for distinguishing subtle differences between live and spoof faces.
}

\begin{table*}[]
\centering
\scriptsize 
\setlength{\tabcolsep}{7pt}
\caption{Comparison of cross-domain face presentation attack detection.
The evaluation metrics are HTER(\%) $\downarrow$ and AUC(\%) $\uparrow$. $\dag$ indicates the transformer-based method. 
\label{tab:cross_testing}}
\resizebox{\columnwidth}{!}{
\begin{tabular}{ccccccccc}
\hline
\multirow{2}{*}{\textbf{Method}}  & \multicolumn{2}{c}{\textbf{{{[}}O,C,I{{]}}$\rightarrow$ M}} & \multicolumn{2}{c}{\textbf{{{[}}O,M,I{{]}}$\rightarrow$ C}} & \multicolumn{2}{c}{\textbf{{{[}}O,C,M{{]}}$\rightarrow$ I}} & \multicolumn{2}{c}{\textbf{{{[}}I,C,M{{]}}$\rightarrow$ O}} \\ \cline{2-9} 
 &   \multicolumn{1}{c}{HTER} & AUC & \multicolumn{1}{c}{HTER} & AUC & \multicolumn{1}{c}{HTER} & AUC & \multicolumn{1}{c}{HTER} & AUC \\ \hline
RAEDFL \cite{huang2022face}   (\textit{ACPR 21}) & \multicolumn{1}{c}{16.67} & 87.93 & \multicolumn{1}{c}{17.78} & 86.11 & \multicolumn{1}{c}{14.64} & 85.64 & \multicolumn{1}{c}{18.06} & 90.04 \\ 
SSAN-M \cite{wang2022domain}   (\textit{CVPR 22}) & \multicolumn{1}{c}{{10.42}} & 94.76 & \multicolumn{1}{c}{16.47} & 90.81 & \multicolumn{1}{c}{{14.00}} & {94.58} & \multicolumn{1}{c}{19.51} & 88.17 \\ 
SSAN-R \cite{wang2022domain}   (\textit{CVPR 22}) & \multicolumn{1}{c}{{6.67}} & {98.75} & \multicolumn{1}{c}{10.00} & {96.67} & \multicolumn{1}{c}{8.88} & {96.79} & \multicolumn{1}{c}{13.72} & {93.63} \\ 
LDCN \cite{huang2022learnable}  (\textit{BMVC 22}) &  \multicolumn{1}{c}{9.29} & 96.86 & \multicolumn{1}{c}{12.00}  & 95.67 & \multicolumn{1}{c}{9.43} & 95.02 & \multicolumn{1}{c}{13.51} & 93.68 \\
ViTranZFAS \cite{george2021effectiveness} $\dag$ (\textit{IJCB 21}) & \multicolumn{1}{c}{10.95} & 95.05 & \multicolumn{1}{c}{14.33} & 92.10 & \multicolumn{1}{c}{ {16.64}} & 85.07 & \multicolumn{1}{c}{15.67} & 89.59 \\ 
TransFAS \cite{wang2022face} $\dag$ (\textit{TBBIS 22}) & \multicolumn{1}{c}{7.08} & 96.69 & \multicolumn{1}{c}{9.81} & 96.13 & \multicolumn{1}{c}{ {10.12}} & 95.53 & \multicolumn{1}{c}{15.52} & 91.10 \\ 
TTN-S \cite{wang2022learning} $\dag$ (\textit{TIFS 22}) & \multicolumn{1}{c}{9.58} & 95.79 & \multicolumn{1}{c}{9.81} & 95.07 & \multicolumn{1}{c}{ {14.15}} & 94.06 & \multicolumn{1}{c}{12.64} & 94.20 \\
SA-FAS \cite{sun2023rethinking}  (\textit{CVPR 23})   & \multicolumn{1}{c}{5.95} & 96.55 & \multicolumn{1}{c}{8.78} & 95.37 & \multicolumn{1}{c}{ {6.58}} & 97.54 & \multicolumn{1}{c}{10.00} & 96.23 \\ 
DFANet \cite{huang2023towards}  (\textit{ICME 23})   & \multicolumn{1}{c}{5.24}         & 97.98    & \multicolumn{1}{c}{8.78}        & 97.03 & \multicolumn{1}{c}{8.21}        & 96.84 & \multicolumn{1}{c}{9.34}       & 96.43 \\ 
UDG-FAS+SSDG   \cite{liu2023towards}  (\textit{ICCV 23})   & \multicolumn{1}{c}{5.95}         & 98.47    & \multicolumn{1}{c}{9.82}        & 96.76 & \multicolumn{1}{c}{5.86}        & 98.62 & \multicolumn{1}{c}{10.97}       & 95.36 \\ 

FDGAL   \cite{huang2023face}  (\textit{PR 23})   & \multicolumn{1}{c}{12.92}         & 93.29    & \multicolumn{1}{c}{17.78}        & 88.10 & \multicolumn{1}{c}{18.75}        & 91.92 & \multicolumn{1}{c}{15.90}       & 90.54 \\ 


\textcolor{red}{HPDR \cite{hu2024rethinking} (\textit{CVPR 24})} & \multicolumn{1}{c}{4.58} & 96.02 & \multicolumn{1}{c}{11.30} & 94.42 & \multicolumn{1}{c}{11.26} & 92.49 & \multicolumn{1}{c}{9.93} & 95.26 \\

     
\textcolor{red}{DFDN  \cite{ma2024dual}  (\textit{PR 24})}   & \multicolumn{1}{c}{5.20}         & 98.39    & \multicolumn{1}{c}{8.00}        & 97.45 & \multicolumn{1}{c}{7.71}        & 95.56 & \multicolumn{1}{c}{11.01}       & 95.22 \\  
\hline
LDCformer $\dag$ &  \multicolumn{1}{c}{\textbf{4.52}} & \textbf{98.77} & \multicolumn{1}{c}{\textbf{2.33}} & \textbf{99.70} & \multicolumn{1}{c}{{\textbf{5.36}}} & {\textbf{98.69}} & \multicolumn{1}{c}{\textbf{6.77}} & \textbf{97.89} \\ 
\hline
\end{tabular} }
\end{table*}

\subsubsection{Cross-Domain Testing} 
We first follow the setting of \cite{shao2019multi} to conduct cross-domain testing by using the model trained on multiple training domains to detect face presentation attacks on one unseen domain. 
Table \ref{tab:cross_testing} shows the detection performance of four cross-domain testing protocols on the datasets \textbf{OULU-NPU}, \textbf{MSU-MFSD}, \textbf{CASIA-MFSD}, and \textbf{Idiap Replay-Attack}. 
\textcolor{red}{
In cross-domain testing scenario, previous FAS methods have employed various techniques to enhance model generalization.}
\textcolor{red}{
For example, in SSAN-M/SSAN-R \cite{wang2022domain}, DFANet \cite{huang2023towards}, and DFDN \cite{ma2024dual}, the authors proposed learning disentangled liveness features to tackle cross-domain issue. 
Similarly, the authors in FDGAL \cite{huang2023face} proposed adopting global attention learning to guide the model in capturing subtle spoofing cues, thereby enhancing its ability to distinguish live faces from spoofed ones.
However, these methods overlook a crucial factor in enhancing model generalization: spoof attacks of the same type often exhibit similar characteristics.
Therefore, by examining these similar spoof characteristics within the same attack type to learn the live-to-spoof transition, LDCformer, guided by the proposed transitional triplet mining, effectively learns domain-invariant and domain-generalized liveness features.}
The results in Table \ref{tab:cross_testing} show that LDCformer outperforms previous methods
and achieves state-of-the-art performance on the four protocols in both metrics HTER and AUC. 
This performance improvement demonstrates the efficacy of LDCformer and the collaborative training strategies.

\begin{table}[]
\centering
\scriptsize 
\setlength{\tabcolsep}{3pt}
\caption{Comparison of limited cross-domain testing on \textbf{{[}M, I{]}$\rightarrow$ C} and \textbf{{[}M, I{]}$\rightarrow$ O}. The evaluation metrics are HTER(\%) $\downarrow$ and AUC(\%) $\uparrow$. 
$\dag$ indicates the transformer-based method. 
\label{tab:limited_cross_type}}
\begin{tabular}{ccccc}
\hline
\multirow{2}{*}{\textbf{Method}} &  \multicolumn{2}{c}{\textbf{{[}M,I{]}$\rightarrow$ C}} & \multicolumn{2}{c}{\textbf{{[}M,I{]}$\rightarrow$ O}} \\ \cline{2-5} 
 &   \multicolumn{1}{c}{HTER} & AUC & \multicolumn{1}{c}{HTER} & AUC \\ \hline
RAEDFL \cite{huang2022face}  (\textit{ACPR 21}) & \multicolumn{1}{c}{31.11} & 72.63 & \multicolumn{1}{c}{29.23} & 74.62 \\ 
ViTranZFAS \cite{george2021effectiveness} $\dag$ (\textit{IJCB 21}) & \multicolumn{1}{c}{23.54} &  {81.07} & \multicolumn{1}{c}{26.09} &  {80.38} \\ 
SSAN-M \cite{wang2022domain}  (\textit{CVPR 22}) & \multicolumn{1}{c}{30.00} & 76.20 & \multicolumn{1}{c}{29.44} & 76.62 \\ 
LDCN \cite{huang2022learnable}  (\textit{BMVC 22}) & \multicolumn{1}{c}{22.22} & 82.87 & \multicolumn{1}{c}{21.54} & 86.06 \\
DFANet \cite{huang2023towards}  (\textit{ICME 23}) & \multicolumn{1}{c}{20.67} & 84.87 & \multicolumn{1}{c}{18.61} & 89.52 \\
 \textcolor{red}{MFAE \cite{zheng2024mfae} $\dag$ (\textit{TIFS 24})} & \multicolumn{1}{c}{20.00} & 88.47 & \multicolumn{1}{c}{18.73} & 88.00 \\

    \textcolor{red}{HPDR \cite{hu2024rethinking} (\textit{CVPR 24})}& \multicolumn{1}{c}{22.22} & 85.54 & \multicolumn{1}{c}{21.07} & 87.53 \\


    \textcolor{red}{GAC-FAS \cite{le2024gradient} (\textit{CVPR 24})} & \multicolumn{1}{c}{16.91} & 88.12 &{17.88} & {89.67} \\
\hline
LDCformer $\dag$ & \multicolumn{1}{c}{\textbf{15.56}} & \textbf{90.09} & \multicolumn{1}{c}{\textbf{17.56}} & \textbf{90.08} \\ \hline
\end{tabular}
\end{table}

\subsubsection{Cross-Domain Testing with Limited Source Domains} Next, we conduct limited cross-domain testing by using the model trained on only two source domains to evaluate the domain generalization ability. 
Among the four datasets, as mentioned in \cite{shao2019multi,huang2022learnable}, there exists a significant domain gap between \textbf{MSU-MFSD} and \textbf{Idiap Replay-Attack}.
Therefore, we use these two datasets as training domains and then conduct the cross-domain testing on the two protocols \textbf{{[}M, I{]}$\rightarrow$ C} and \textbf{{[}M, I{]}$\rightarrow$ O}. 
\textcolor{red}{
Table~\ref{tab:limited_cross_type} shows that FAS models encounter increased challenges as the amount of training source domains decreases.
However, with the proposed self-challenging supervision generating more challenging mixed data, LDCformer can still effectively learn discriminative features, even under limited source domain conditions.} 
The results in Table \ref{tab:limited_cross_type} show that LDCformer significantly outperforms all the other methods. 
These results demonstrate the superior generalization ability of LDCformer even when training on a small number of source domains.

\begin{table*}[]
\footnotesize 
\setlength{\tabcolsep}{5pt}
\caption{ Comparison of cross-domain testing on ``unseen’’ attack types. 
The evaluation metrics are HTER(\%) $\downarrow$ and AUC(\%) $\uparrow$.}
\centering
\resizebox{\columnwidth}{!}{
\begin{tabular}{ccccccc}
\hline
\multirow{2}{*}{\textbf{Method}} &  \multicolumn{2}{c}{\textbf{3DMAD} \cite{erdogmus2014spoofing}} & \multicolumn{2}{c}{\textbf{HKBU-MARs} \cite{liu20163d}} & \multicolumn{2}{c}{\textbf{CASIA-3DMask}\cite{yu2020fas}} \\ \cline{2-7} 
 & \multicolumn{1}{c}{HTER} & {AUC} & \multicolumn{1}{c}{HTER} & AUC & \multicolumn{1}{c}{HTER} & AUC \\ \hline
Auxiliary \cite{liu2018learning}  (\textit{CVPR 18}) & \multicolumn{1}{c}{0.29} & 99.04 & \multicolumn{1}{c}{14.64} & 88.32 & \multicolumn{1}{c}{37.28} & 53.14 \\
\textcolor{red}{BASN \cite{kim2019basn} (\textit{ICCVW 19}) }& \multicolumn{1}{c}{0.55} & 98.10  & \multicolumn{1}{c}{27.09}  & 79.83 & \multicolumn{1}{c}{39.85} &  57.24\\  
NAS \cite{yu2020fas} (\textit{TPAMI 20}) & \multicolumn{1}{c}{0.22} & 99.31 & \multicolumn{1}{c}{15.13} & 88.91 & \multicolumn{1}{c}{37.68} & 72.83 \\  
RFMeta \cite{shao2020regularized} (\textit{AAAI 20}) & \multicolumn{1}{c}{0.20} & 99.50 & \multicolumn{1}{c}{20.39} & 84.30 & \multicolumn{1}{c}{32.29} & 72.96  \\
RAEDFL \cite{huang2022face} (\textit{ACPR 21}) & \multicolumn{1}{c}{0.41}  & 99.38  & \multicolumn{1}{c}{16.89} & 86.26 & \multicolumn{1}{c}{38.70}  & 65.37  \\
\textcolor{red} {ViTranZFAS \cite{george2021effectiveness} $\dag$ (\textit{IJCB 21}) } & \multicolumn{1}{c}{1.19}  & 99.77  & \multicolumn{1}{c}{15.29} & 89.86 & \multicolumn{1}{c}{36.00}  & 66.36  \\
\textcolor{red} {PatchNet \cite{wang2022patchnet} (\textit{CVPR 22}) } & \multicolumn{1}{c}{0.60} & 99.95  & \multicolumn{1}{c}{25.74} & 73.20  &  \multicolumn{1}{c}{{48.60}} &  {52.17}  \\
LDCN \cite{huang2022learnable} (\textit{BMVC 22}) & \multicolumn{1}{c}{1.49} & 99.91  & \multicolumn{1}{c}{8.75} & 95.60 &  \multicolumn{1}{c}{{33.54}} &  {60.44}  \\
DFANet \cite{huang2023towards} (\textit{ICME 23}) & \multicolumn{1}{c}{1.09} & 99.13  & \multicolumn{1}{c}{10.64} & 92.31 &  \multicolumn{1}{c}{{33.10}} &  {69.28}  \\
\textcolor{red}{GAC-FAS \cite{le2024gradient} (\textit{CVPR 24})} & \multicolumn{1}{c}{1.19} &  99.97 & \multicolumn{1}{c}{18.39} & 87.00 &  \multicolumn{1}{c}{37.05} & 68.69\\
\hline
LDCformer &  \multicolumn{1}{c}{\textbf{0.17}} & \textbf{99.99} & \multicolumn{1}{c}{\textbf{8.60}} & \textbf{97.19} & \multicolumn{1}{c}{\textbf{30.83}} & {\textbf{75.57}} \\ 
\hline
\end{tabular} }
\label{tab:cross_domain_inter_type}
\end{table*}

\subsubsection{ Cross-Domain Testing on ``Unseen’’ Attack Types}
In Table \ref{tab:cross_domain_inter_type}, we evaluate domain generalization ability in a challenging setting by detecting ``unseen attack types’’ on unseen domains. We follow the ``Cross-Dataset Cross-Type Protocol’’ in \cite{yu2020fas} to conduct this experiment by using \textbf{OULU-NPU} and \textbf{SiW} for training and then testing on \textbf{3DMAD}, \textbf{HKBU-MARs}, and \textbf{CASIA-3DMask}. Note that, the methods \cite{liu2018learning,shao2020regularized,huang2022face}, which adopted depth-based auxiliary supervision, have severely degraded performance when testing on \textbf{HKBU-MARs}, because facial depths offer little discriminative information for detecting the 3D mask attacks in \textbf{HKBU-MARs}. 
\textcolor{red}{
The method \cite{kim2019basn}, which adopted image reflection-based auxiliary supervision, has a poor performance on \textbf{CASIA-3DMask} because the no-reflection assumption for live faces is no longer valid in the outdoor lighting condition of \textbf{CASIA-3DMask}.
In contrast, LDCformer, guided by the proposed transitional triplet mining, benefits from clustering live faces, enabling the model to learn what constitutes a live face to better tackle unseen attacks.
Under the collaborative supervision of three training strategies, our method LDCfomer substantially outperforms other methods, and demonstrates its excellent domain generalization ability in detecting unseen attack types.
}

\subsubsection{\textcolor{red}{Cross-Domain Testing on ``Partial’’ Attack Types}} 

\textcolor{red}{
In Table~\ref{tab:partial_attacks}, we conduct cross-domain ``partial’’ attack testing on \textbf{PADISI-Face} \cite{rostami2021detection}.
In particular, we train the FAS models using full-face spoof attacks and half of the live images from \textbf{PADISI-Face} \cite{rostami2021detection} and then test on partial spoof attacks and the remaining live images from the same dataset.
Since previous FAS methods primarily focus on learning liveness features to address full-face spoof attacks, they often exhibit poor performance in detecting partial spoof attacks.
In contrast, with the proposed self-challenging supervision, LDCformer significantly benefits from the generated mixed challenging data, enabling it to effectively distinguish between partial spoof attacks and live faces.
}

\begin{table}[t]
\centering
\scriptsize
\color{red}
\setlength{\tabcolsep}{2pt}
\caption{
Comparison of cross-domain partial attack detection on \textbf{PADISI-Face} \cite{rostami2021detection}.
The evaluation metrics are APCER(\%) $\downarrow$, BPCER(\%) $\downarrow$, and ACER(\%) $\downarrow$.  
\label{tab:partial_attacks}
}
\begin{tabular}{c|cc|cc}
\hline
\multirow{2}{*}{Method} & \multicolumn{2}{c|}{Funny eye attacks}          & \multicolumn{2}{c}{Paper glasses attacks
}           \\ \cline{2-5} 
                        & \multicolumn{1}{c|}{HTER} & AUC & \multicolumn{1}{c|}{HTER} & AUC \\ \hline
LDCN \cite{huang2022learnable} (\textit{BMVC 22}) & \multicolumn{1}{c|}{22.08} & 83.32 & \multicolumn{1}{c|}{16.30} & 86.09 \\ \hline
ViTranZFAS \cite{george2021effectiveness} $\dag$ (\textit{IJCB 21}) & \multicolumn{1}{c|}{21.50} & 85.03 & \multicolumn{1}{c|}{14.14} & 87.12 \\ \hline
SSAN \cite{wang2022domain} (\textit{CVPR 22}) & \multicolumn{1}{c|}{17.16} & 88.73 & \multicolumn{1}{c|}{7.86} & 96.15 \\ \hline
GAC-FAS \cite{le2024gradient} (\textit{CVPR 24})& \multicolumn{1}{c|}{12.94} & 92.97 & \multicolumn{1}{c|}{7.22} & 97.31 \\ \hline
LDCformer & \multicolumn{1}{c|}{8.63} & 96.23 & \multicolumn{1}{c|}{2.35} & 99.33 \\ \hline
\end{tabular}
\end{table}

\subsection{\textcolor{red}{Limitation}}
\textcolor{red}{
In  Table \ref{tab:ablation_study_limitation}, we compare using different backbones to train LDCformer under the same loss terms  $\mathcal{L}_{ls} + \mathcal{L}_{dual}+\mathcal{L}_{sc}+\mathcal{L}_{tran-trip}$ and using different metrics, including the number of parameters (\#param.), FLOPs, and FPS, to measure the model size, the computational complexity, and the inference throughput on the protocol \textbf{{[}I,C,M{]} $\rightarrow$ O}.
The batch size is set to 18 for this experiment.}
\textcolor{red}{
While the proposed training strategies effectively guide LDCformer (with the standard ViT) to achieve promising performance, both model training and inference require additional computational resources. Furthermore, as the model size decreases, inference speed improves, but performance declines to some extent. Therefore, there remains a need to find an optimal balance between effectiveness and efficiency for LDCformer in practical real-world applications.
}


\begin{table}[t]
\color{red}
\centering
\footnotesize
\caption{
\textcolor{red}{
Ablation study on \textbf{{[}I,C,M{]} $\rightarrow$ O}, using different backbones and the same loss for model training. 
The evaluation metrics are HTER(\%) $\downarrow$ and AUC(\%) $\uparrow$.}
\label{tab:ablation_study_limitation}
}
\begin{tabular}{c|cc|c|c|c}
\hline
\multirow{2}{*}{Backbone} & \multicolumn{2}{c|}{\textbf{{[}I,C,M{]} $\rightarrow$ O}} & \multirow{2}{*}{\#param.} & \multirow{2}{*}{FLOPs} & \multirow{2}{*}{FPS} \\ \cline{2-3}
                          & \multicolumn{1}{c|}{HTER} & AUC & & & \\ \hline
Mobile ViT  \cite{dosovitskiy2021an}                & \multicolumn{1}{c|}{10.32}     &   95.92  & 5.63M & 33.08G & 76.71 \\ \hline
LDCformer  & \multicolumn{1}{c|}{6.77}& 97.89 & 87.57M & 303.91G & 32.73 \\ \hline
ResNet18\cite{he2016deep} & \multicolumn{1}{c|}{11.96} & 93.95 & 12.51M & 33.99G & 53.72 \\ \hline
ResNet34\cite{he2016deep} & \multicolumn{1}{c|}{11.77} & 94.82 & 22.62M & 67.38G & 48.85 \\ \hline
\end{tabular}
\end{table}

\section{Conclusion}  
\textcolor{red}{
In this paper, we propose three novel training strategies to enhance the learning of our previously proposed Learnable Descriptive Convolutional Vision Transformer (LDCformer).
First, to address the lack of fine-grained supervision issue,  we propose dual-attention supervision to guide LDCformer in learning fine-grained liveness features. 
Next, to detect subtle partial spoof attacks, we propose an effective self-challenging supervision method to simulate these attacks to guide LDCformer for learning discriminative liveness features.
Moreover, to tackle the cross-domain issue, we propose transitional triplet mining to further encourage LDCformer on enhancing the domain generalization ability of the learned features.
Our comprehensive ablation study, along with extensive intra-domain and cross-domain testing on face anti-spoofing benchmarks, demonstrates significant performance improvements and validates the effectiveness of the proposed three training strategies.
Furthermore, we show that these strategies not only benefit ViT architectures but can also be generalized to conventional CNN architectures.   
Nevertheless, while LDCformer achieves superior performance, it requires more computational resources and computation time compared to CNN models. Therefore, further study is needed to find an optimal balance between effectiveness and efficiency for practical real-world applications.
For future work, there are two promising directions for extending LDCformer in FAS.
First, our current implementation of LDCformer focuses on single-modal input data - RGB data.
However, because RGB data are much sensitive to lighting variations, multi-modal FAS methods become increasingly important for detecting spoof attacks in complex environments. 
Second, since off-line training data cannot cover all possible spoof attacks, online learning through test-time adaptation (TTA) \cite{huang2023test} or one-class face anti-spoofing (OC-FAS) \cite{huang2024one}, which focuses on learning liveness features from live faces alone, offers a more promising direction for future research.
}

\bibliographystyle{elsarticle-num-names} 
\bibliography{egbib}





\end{document}